\documentclass{article}

\usepackage{microtype}
\usepackage{graphicx}
\usepackage{subfigure}
\usepackage{booktabs} 

\usepackage{url}
\usepackage{hyperref}

\usepackage[accepted]{icml2025}

\usepackage{amsmath}
\usepackage{amssymb}
\usepackage{mathtools}
\usepackage{amsthm}

\usepackage[capitalize,noabbrev]{cleveref}

\theoremstyle{plain}

\theoremstyle{definition}

\theoremstyle{remark}

\usepackage[textsize=tiny]{todonotes}

\icmltitlerunning{Yield Curve Forecasting with ML}

\usepackage{soul}
\usepackage{appendix}
\usepackage{bibspacing}

\begin{document}

\twocolumn[
\icmltitle{\large Yield Curve Forecasting using Machine Learning and Econometrics: A Comparative Analysis}



\icmlsetsymbol{equal}{*}

\begin{icmlauthorlist}
\icmlauthor{Aman Singh}{aff1}
\icmlauthor{Tokunbo Ogunfunmi}{aff1}
\icmlauthor{Sanjiv Das}{aff2}
\end{icmlauthorlist}

\icmlaffiliation{aff1}{Department of Electrical and Computer Engineering, Santa Clara University, USA}
\icmlaffiliation{aff2}{School of Business, Santa Clara University, USA}

\icmlcorrespondingauthor{Aman Singh}{a2singh@scu.edu}

\icmlkeywords{yield curve, bonds, forecasting, time series, transformers, ARIMA, RNNs, machine learning, deep learning}

\vskip 0.3in
]



\printAffiliationsAndNotice{}  

\begin{abstract}
While machine learning has revolutionized many fields such as natural language processing (NLP) and computer vision, its impact on time-series forecasting is still widely disputed, especially in the finance domain. This paper compares forecasting performance on U.S. Treasury yield curve data across econometrics/time-series analysis, classical machine learning, and deep learning methods, using daily data over 47 years. The Treasury yield curve is important because it is widely used by every participant in the bond markets, which are larger than equity markets. We examine a variety of methods that have not been tested on yield curve forecasting, especially deep learning algorithms. The algorithms include the Autoregressive Integrated Moving Average (ARIMA) model and its extensions, naive benchmarks, ensemble methods, Recurrent Neural Networks (RNNs), and multiple transformers built for forecasting. ARIMA and naive econometric models outperform other models overall, except in one time block. Of the machine learning methods, TimeGPT, LGBM and RNNs perform the best. Furthermore, the paper explores whether stationary or nonstationary data are more appropriate as input to deep learning models.

\end{abstract}

\section{Introduction}

\subsection{Motivation}
Machine learning, especially deep learning, has revolutionized fields such as natural language processing (NLP), computer vision, and generative modeling. RNNs/LSTMs dominated NLP up until Transformers were invented in \cite{Vaswani2017}.
 Eventually, Transformers were scaled up and trained on huge corpora of text data in a self-supervised manner, creating large language models (LLMs). Convolutional Neural Networks (CNNs) have dominated computer vision applications for over a decade. For generative vision tasks, Variational Autoencoders (VAEs) and Generative Adversarial Networks (GANs) were initially dominant \cite{Singh2022,Goodfellow2014GAN,KingmaWelling2014}, until diffusion models were created. 

LLMs and diffusion models are examples of foundation models. A ``foundation model'' is a very large machine learning model ``that is trained on broad data (generally using self-supervision at scale) that can be adapted (e.g., fine-tuned) to a wide range of downstream tasks'' \cite{Bommasani2021}.
In the context of time-series forecasting, machine learning and deep learning have not been as dominant; there is controversy over whether they are more effective than traditional or even naive methods.

The Makridakis Competitions are a series of time-series forecasting competitions whose purpose is to compare the accuracy of different forecasting algorithms; they are also known as the M competitions. The 2018-2019 M4 Competition employed an extensive dataset comprising 100,000 time-series to determine the most precise forecasting techniques tailored for various prediction categories
\cite{makridakis2020m4}. What M4 found was that pure machine learning/neural network models performed poorly and that forecast combinations won 12 of the top 17 models.
However, the winner was Exponential Smoothing-Recurrent Neural Networks (ES-RNN), a hybrid of statistical and deep learning
\cite{smyl2020}.
In second place was a forecast combination of seven statistical methods and one machine learning approach. The weighting for the averaging was determined by a machine learning algorithm. Since the release of ES-RNN, multiple deep learning models such as NBEATS, NHITS, and DeepAR have been benchmarked on the M4 data, beating ES-RNN.

Multiple transformer models have been created with the aim of tackling forecasting, such as Temporal Fusion Transformer (TFT) \cite{lim2021}, Fedformer \cite{fedformer}, Autoformer \cite{autoformer}, Informer \cite{zhou2021}, PatchTST \cite{nie2023a}, Pyraformer \cite{Liu2022Pyraformer}, and iTransformer \cite{liu2024itransformerinvertedtransformerseffective}. Transformer methods bear comparison with traditional time-series econometric approaches for forecasting interest rates, which is the focus of this paper. 

In \citet{Zeng2023} FEDformer, Autoformer, Informer, and Pyraformer are tested on forecasting benchmark datasets, finding that they generally did not perform as well as a simple Linear Model called DLinear, with respect to the MSE and MAE metrics. It is suggested that because the self-attention mechanism is permutation-invariant, there is temporal information loss. \citet{BlogPost2023} offers a response with a couple of experiments using the MASE metric to evaluate the results. The first experiment compares an Autoformer (univariate) with DLinear on three benchmark datasets finds that Autoformer does better. The second experiment compares the Vanilla Transformer (univariate), the Vanilla	Transformer (multivariate), the	Informer (univariate), the Informer (multivariate), the Autoformer (univariate), and 	DLinear; this finds that the Vanilla Transformer does better. More transformers such as PatchTST and Temporal Fusion Transformers have been released since the paper. For transformers there doesn't seem to be an agreement on the best performance.

In 2023, {\tt Nixtla}\footnote{\url{https://www.nixtla.io}} released a transformer-based foundational model for forecasting called TimeGPT \cite{Garza2024} that shows good performance versus other major methods on 300,000 time-series data sets from multiple domains. More foundational models for forecasting have been released since, such as TimeFM \cite{Sen2024}, Chronos \cite{Ansari2024}, Time-LLM \cite{Jin2024}, and MOIRAI \cite{Woo2024}. 

Hybrid methods and deep learning methods can show promise in forecasting, though it does not imply good performance for every forecasting problem. In this paper we focus on the finance industry, which commonly forecasts volatility, price, returns, or the direction of financial assets, mostly in the realm of equities. Less emphasis has been placed on ML forecasting for Treasury constant maturity yield curves, especially using deep learning methods. Yield curves are important because they are essential for bond trading and can be predictive of recessions. 

Another area of interest is how well deep learning algorithms can deal with nonstationary data. Often times, forecasting algorithms need to difference the input data to make it stationary or require the input data to be made stationary before using it. It is hypothesized that deep learning algorithms can deal with nonstationary data directly. See Appendix \ref{stationarity} for more details on stationarity. We consider both types of input data, with and without treatments for nonstationarity. 

Before summarizing the contributions of the paper, we review the extant literature. 

\subsection{Literature Review}

Most of the yield curve forecasting literature focuses on {\it monthly} yield curve forecasting. We use {\it daily} data. 

Traditional time-series/econometrics models for monthly yield curve forecasting include  AutoRegressive Models (AR models), Vector AutoRegressive  (VAR) Models, Bayesian VAR models, and Factor Augmented VAR  models \cite{caldeira2013, moench2006}. In \citet{caldeira2013}, the authors used forecast combinations. In \citet{SwansonXiong2018}, the authors compare the Dynamic Nelson-Siegel model \cite{NelsonSiegel1987} with AR, VAR, and Principal Component Analysis (PCA) models. Comparisons use big data models vs. small data models on end-of-month yield curve data.

Machine learning has been used for monthly yield curve forecasting for 1 step ahead prediction. In \citet{sambasivan2017}, for US Treasury Constant Maturities, the Dynamic Gaussian Process approach was compared with  the Nelson-Siegel model and VAR; the authors did not use any other features. They found that the Dynamic Gaussian Process and VAR performed better than the N-S model. For the short term region of the yield curve, the VAR performs better than the Dynamic Gaussian Process, but the Dynamic Gaussian Process performs better for the medium-term and long-term maturities. In \citet{reinicke2019}, the authors compared Functional PCA Regression with the Dynamic Nelson-Siegel (with AR/VAR) and the Dynamic Gaussian Process on the US and German yield curves. They used an expanding window approach evaluated with the RMSE metric. They also explored how the size of the training window affects the forecasting performance. 

In \citet{Zhang2022}, the authors attempt to unify machine learning based factors models with no arbitrage pricing theory to predict movements in China's exchange-based treasury market yield curves. In \citet{Rahimi2020}, the authors compare Random Forests, Functional Non-parametric and Dynamic Nelson-Siegel models. They find that the Random Forest model is superior in  forecasting the short end of the yield curve and the Dynamic Nelson-Siegel model is superior in predicting yields of bonds with long term to maturity; they find that using macroeconomic variables is also important. \citet{oosterlaken2020} aims to recreate and extend the work of \citet{SwansonXiong2018}. They compare traditional methods (AR models, VAR models, Dynamic Nelson-Siegel models, and  Principal Component Analysis) vs. machine learning (Regression Trees, Random Forests, Gradient Boosted Trees, and Gaussian Process Regression). They used monthly data, multiple forecast horizons, and find that machine learning models do not perform better than AR(1). 

\citet{bianchi2020} use machine learning to predict excess Treasury bond returns. They used PCR, PLS, Penalized Regressions, Regression Trees,  Neural Networks (yields only) and Neural Networks (yield+macroeconomic variables). They aimed to see if improvements in predictive accuracy led to enhanced investment performance compared to the scenario with no predictability. They find that ``nonlinear models'' such as trees and neural networks identify substantial, statistically significant variations in bond returns. The predictions generated by these techniques result in substantial economic gains on the test set. They also find that with neural networks, adding macroeconomic variables gives higher returns than just using yield data.


In  \cite{hoogteijling2020}, the authors build upon \cite{bianchi2020}, but point out two flaws. The first one is that they used nonstationary inputs instead of stationary inputs. Second, it is often the case that macroeconomic variables for April might be released until 2 months later; they did not account for this and used variables that would not be available  at the time of the prediction.

In  \citet{Chong2021}, the authors have compared the US Treasury Constant Maturity yields (monthly basis) in the time period January 2017 to April 2021. They used AdaBoost, k-NN, Linear Regression, MLP, Random Forest, SVM, AR models, and VAR models. They also use a set of economic features as exogenous variables. SVM outperformed the rest for forecasting 20 Year and 30 Year bond yield while ARIMA outperformed other models for the shorter-ends of the yield curve but the performance of ARIMA dropped significantly if the number of forecast steps are increased. Overall, the machine learning approach is generally preferred due to its higher accuracy for long-term yield forecasting.

\subsection{Contributions}

While these studies have provided valuable insights into the application of various models for yield curve forecasting,  this paper also aims to address several additional aspects that have been unexplored/underexplored. Specifically, the main goals are:
(1) Examine the performance of machine learning (classical machine learning and deep learning) versus traditional econometrics/time-series methods.
(2) Check performance on multiple forecast horizons and throughout different time periods.
(3) Examine whether expanding versus sliding window training/evaluation works better.
(4) Compare stationary versus nonstationary inputs for deep learning algorithms.
This paper is more comprehensive in the number of methods covered, including deep learning models such as small transformers and foundation models that have not been tested on yield curve prediction. We use appropriate benchmarks to compare the methods. Recent papers also typically use monthly data; we use daily data. There is sparse literature on stationary versus nonstationary data inputs for deep learning, which we complement in this research. 
The main findings are that ARIMA and the Naive Method outperform all other models overall, except in one time block. Of the machine learning methods, TimeGPT, LGBM, and RNNs perform the best.

\section{Methodology}
In this subsection, we describe the data being used, the methodology and the experimental setup.

\subsection{Data}
\label{sec:data}

\begin{figure*}
\centering
\includegraphics[scale=0.27]{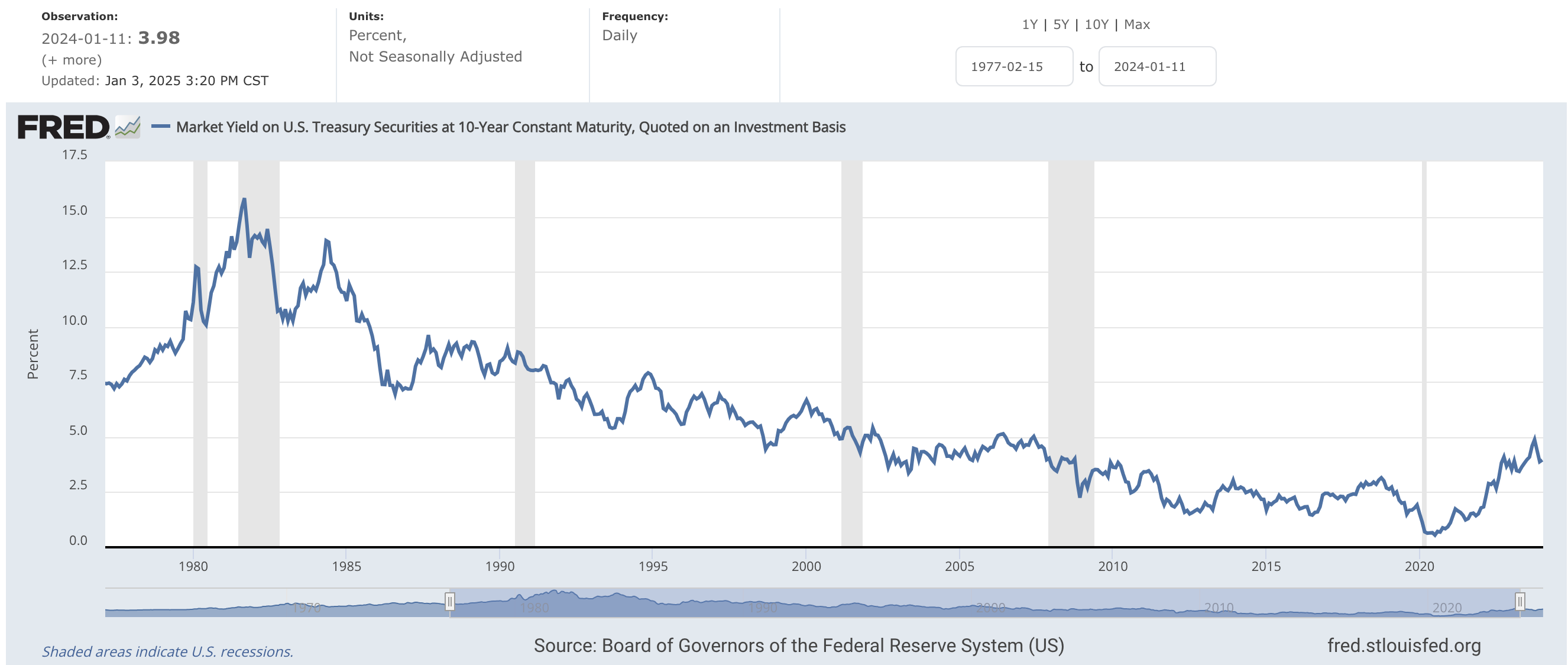}
\includegraphics[scale=0.32]{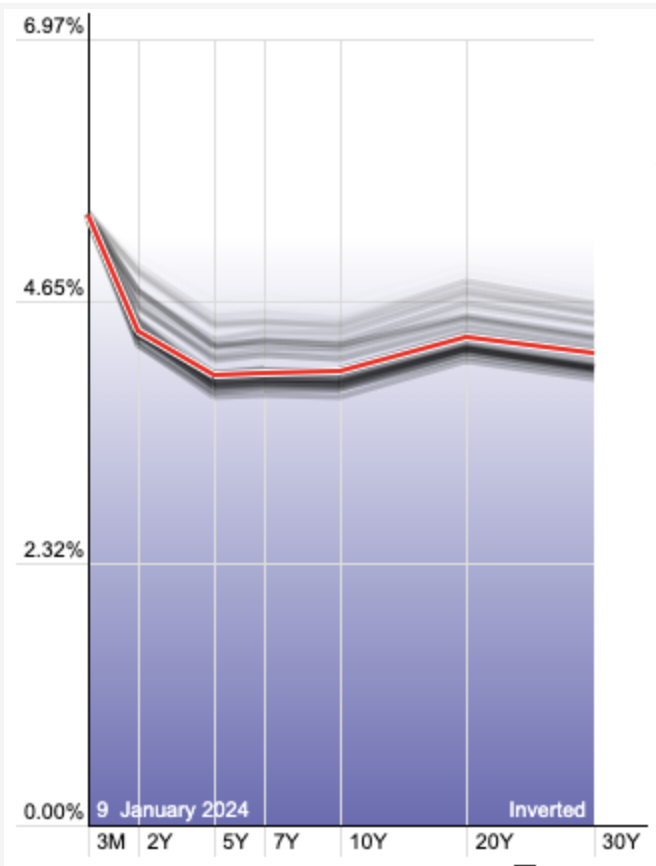}
\caption{\label{fig:dgs10} \scriptsize Left: Time series of the 10-year constant maturity interest rate (DGS10) over the 47-year sample period used in the paper. The page on FRED for this data is at \url{https://fred.stlouisfed.org/series/DGS10}. Useful information about the Treasury's interest rate statistics is at \url{https://home.treasury.gov/policy-issues/financing-the-government/interest-rate-statistics}. The methodology for constructing these yields is provided here: \url{https://home.treasury.gov/policy-issues/financing-the-government/interest-rate-statistics/treasury-yield-curve-methodology}. Right: The red line shows the yield curve on January 9, 2024, around the end of the sample period. \url{https://stockcharts.com/freecharts/yieldcurve.php}. The grey lines show the yield curves for the preceding two weeks. The 10-year yield on this date is around 4.5\%, and this corresponds to the level seen for DGS10 on the left side plot. 
}
\end{figure*}

We use constant maturity yields for six tenors of 1, 2, 5, 7, 10, and 30 years from the FRED database.\footnote{\tiny \url{https://fred.stlouisfed.org}} This is high-quality data from the Federal Reserve. The corresponding tickers are DGS1, DGS2, DGS5, DGS7, DGS10, and DGS30 form the yield curves from 1977-02-15 to 2024-01-11, i.e., a multivariate time series spanning daily data for 47 years, including six recessions. 

The {\it yield} on a bond of maturity $\tau$ years is the annual rate of return if the bond is bought and held to its maturity (this is also known as the bond's {\it yield-to-maturity}, or YTM). On any given day, the curve connecting the YTMs of bonds with different maturities is called the {\it yield curve}. 
For example, the time series for the yield of the current 10-year maturity bond (DGS10)  and the yield curve (for all bonds) on a single day are shown in Figure \ref{fig:dgs10}. We have chosen six points on the yield curve to forecast---these broadly capture the structural change in the yield curve over time for a full spectrum of maturities. The US Treasury typically uses a variant of cubic splines interpolation to fit a smooth curve through the yields to portray the yield curve for each day.\footnote{{\tiny \url{https://home.treasury.gov/policy-issues/financing-the-government/interest-rate-statistics/treasury-yield-curve-methodology}}} This methodology was updated in mid-2021.\footnote{{\tiny \url{https://home.treasury.gov/quasi-cubic-hermite-spline}\\\url{-treasury-yield-curve-methodology}}} 

Treasury bills (short maturity) and bonds (medium to long maturities) are traded continuously and their prices are observable. They provide cashflows in the form of interest rate payments every six months. The YTM is the single rate of return at which the future cashflows are discounted such that the sum of these discounted cashflows equals the current trading price of the bond. Thus, yields are mathematically implicit from the prices of bonds and their interest payments. Since Treasury yields are used not only for pricing Treasury securities but all bonds in the market, forecasting yields is a critical information generation activity for all bond market participants, irrespective of the type of bonds they trade. Assessing the extant yield curve forecasting techniques and modern ML/DL techniques for improvements is of innate interest to bond market traders, risk managers, regulators, and even participants in equity and commodity markets, as yields enter models in all these markets.  The US bond market value in 2022 was \$51 TN\footnote{\tiny \url{https://www.weforum.org/stories/2023/04/ranked-the-largest-bond-markets-in-the-world/}},  larger than the US equity markets (\$40 TN).\footnote{\tiny \url{https://www.ceicdata.com/en/indicator/united-states/market-capitalization}} Wall Street firms have special employees known as ``Fed-watchers''. By forecasting yield curves, Fed watchers aim to anticipate changes in monetary policy, economic conditions, and potential market movements, providing valuable insights for bond market participants and investors across various asset classes.

\subsection{Models and Setup}

We compare the performance of machine learning methods with traditional time-series/econometrics methods.
The traditional time-series/econometrics methods include ARIMA, DHR-ARIMA, VAR, and VECM \cite{ShumwayStoffer2006,Chen2022}. Within machine learning, two groups -- the deep learning group and the classical machine learning group. The former group includes the RNN, the Vanilla Transformer, the Informer, PatchTST, TFT, TimeGPT, DeepAR, NHITS, and NBEATS \cite{Salinas2020,Challu2023,Oreshkin2019}. The Vanilla Transformer and the RNN are specifically from Nixtla's implementation. The latter group includes Random Forests, XGBoost, and LGBM. All the methods are compared against 3 naive benchmarks: the Mean Forecast, the Naive Forecast, and the Naive Seasonal Forecast.

The Naive Forecast uses the last value from the training window and repeats this value for the entire forecast horizon. This is also known as the random walk benchmark. The Naive Seasonal Forecast uses values from the last season and repeats these values for the forecasts. The Mean Forecast uses the mean of the training window and uses it for the entire forecast horizon.

Transformer-based forecasting models like TimeGPT adapt the attention mechanism to capture temporal dependencies over long horizons. Unlike ARIMA or VAR models, which rely on explicitly specified lags and assumptions of stationarity, transformers learn relationships across time series and scale without pre-specification. Self-attention lets the model flexibly determine which parts of the historical series are most relevant for forecasting future values. While all transformer-based models in this paper share the core architecture of attention mechanisms, TimeGPT differs significantly from the smaller forecasting transformers like Informer, PatchTST, and the Vanilla Transformer. TimeGPT is a pretrained foundation model, trained on hundreds of thousands of time series from diverse domains using self-supervision. It is designed to generalize across domains with little or no fine-tuning. In contrast, smaller transformers like Informer and PatchTST are task-specific models, trained from scratch on the target dataset (in this case, Treasury yield curves). They do not benefit from pretraining, and their performance heavily depends on the structure of the local dataset and the training regime. As a result, TimeGPT aims to leverage transfer learning and scale, while small transformers rely solely on learning from scratch, making them more flexible in theory but often less robust in practice when data is limited or noisy.

For ARIMA, we use {\tt autoarima} for hyperparameter selection, which uses AIC as the selection criteria. For VAR and VECM, we also use information criteria-based order selection. For the deep learning models, we select hyperparameters using cross validation (CV).

We use the following metrics: RMSE (root mean-squared error) and MAPE (mean absolute percentage error). Whereas RMSE is not normalized for the level of the interest rate and consequently the size of the prediction error, MAPE is, thus the latter supports comparison of prediction accuracy over time. 

For training and evaluating the performance of the algorithms, we use two setups: sliding window CV and expanding window CV; more details on these approaches are listed in Appendix \ref{data_partitioning}. These setups avoid any forward looking data, which prevents data leakage. Further, the only pre-trained model we used is TimeGPT, which is trained on many time series, so it is possible that it has seen the yields before. However, we do not pass the dates of the time series to the model, thus making it harder for it to use its memory for prediction. We also did not undertake any fine-tuning and used the model in zero-shot mode, thus reducing any chance of leakage. Given that it does not beat the econometric models suggest that information leakage is not an issue. 

With the sliding window CV, the window  size is 1044 business days, which is about 4 years. This window is used to train the forecasting algorithm, which then outputs a forecast over the next 15 business days, which will be evaluated by metrics.

The window will then slide by 5 business days. This new window of data will be used to re-estimate the model parameters and then will output a forecast that will be evaluated. We then move to the next window and repeats the process, until  we no longer have enough data for a test window with 15 days.

With the expanding window CV, the initial window is 1044 days, and it expands by 5 business days each window instead of sliding.

Within the sliding window CV and expanding window CV setups, we  also split the data into what we will call  hyperparameter time blocks to account for structural shifts in the economy over time. Table \ref{time_blocks} shows these blocks for the expanding window evaluation and the sliding window evaluation. These can be viewed as a set of contiguous training windows that share the same hyperparameters. For an example, Block 0 starts with  a window containing data from 1977-02-15 to 1981-02-13; we train the model, forecast, and evaluate with metrics. Then we move to the next window and repeat the process. When the training window becomes the window that contains data from 1982-02-16  to 1986-02-14, this would be the last window of  hyperparameter time block 0. After it is finished, the model would then switch to a different time block with a new set of hyperparameters. 





\begin{table*}[t]
\centering
\small
\caption{\label{time_blocks} \small The different time blocks for the sliding window.} 
\begin{tabular}{|c|ll|ll|} \toprule
 & \multicolumn{2}{c}{Sliding Windows} & \multicolumn{2}{c}{Expanding Windows} \\
\hline \text { Block } & \text { Start Window } & \text { End Window } & \text { Start Window } & \text { End Window } \\ 
 \midrule
0 & 1977-02-15: 1981-02-13 & 1982-02-16 : 1986-02-14 & 1977-02-15 : 1981-02-13 &1977-02-15 : 1986-02-14  \\
 1 & 1982-02-23 : 1986-02-21 & 1987-02-24 : 1991-02-22 & 1977-02-15 : 1986-02-21&1977-02-15 : 1991-02-22\\
 2 & 1987-03-03 : 1991-03-01 &1992-03-03 : 1996-03-01 & 1977-02-15 : 1991-03-01&1977-02-15 : 1996-03-01 \\
 3 & 1992-03-10 : 1996-03-08 &1997-03-11 :  2001-03-09 & 1977-02-15 : 1996-03-08&1977-02-15 : 2001-03-09\\
 4 & 1997-03-18 : 2001-03-16 &2002-03-12 : 2006-03-10 & 1977-02-15 : 2001-03-16&1977-02-15 : 2006-03-10\\
 5 & 2002-03-19 : 2006-03-17 & 2007-03-20 : 2011-03-18 & 1977-02-15 : 2006-03-17&1977-02-15 : 2011-03-18  \\
 6 &2007-03-27 : 2011-03-25 & 2012-03-27 : 2016-03-25 & 1977-02-15 : 2011-03-25&1977-02-15 : 2016-03-25\\
 7 & 2012-04-03 : 2016-04-01 & 2017-04-04 : 2021-04-02  &1977-02-15 : 2016-04-01&1977-02-15 : 2021-04-02 \\
 8 & 2017-04-11 : 2021-04-09 & 2019-12-17 : 2023-12-15 &1977-02-15 : 2021-04-09 & 1977-02-15 : 2023-12-15 \\
\bottomrule
\end{tabular}
\end{table*}


Hyperparameter selection is done using the first training window in each hyperparameter time block and then used for all windows in the time block. 
For each hyperparameter time block, we repeat the process of hyperparameter selection, training, testing, and sliding/expanding the window.  
Once the last time block is completed, we average the RMSE and MAPE score from each window to get our overall MAPE and RMSE. The entire experiment is also repeated with forecast horizons of 20 and 60 business days in addition to the 15 day forecast horizon. Results are stable across forecast horizons.

\section{Results}
\label{results}

We will first go over the deep learning methods, exploring how each method did for sliding stationary, sliding nonstationary, expanding stationary, and expanding nonstationary configurations (stationary inputs are data that has been treated for nonstationarity). Next, we cover traditional statistics/econometrics and classic machine learning methods, with sliding window  and expanding window configurations. The best configurations for each model appear on what we will call the {\it final model performance table}. 

\subsection{Analysis of Deep Learning Methods}

Table \ref{deeplearning_metrics} shows the error evaluation results of the deep learning methods.  
Table \ref{tab:var1} in Appendix \ref{std_tables}  shows the standard deviation of these results. And there's retard
Of all the deep learning  methods for the overall average, TimeGPT has the best performance for both metrics, RMSE and MAPE. If we exclude TimeGPT, RNN sliding stationary does the best for both metrics. The other seven deep learning methods are not as good as the top two. However, TimeGPT is a foundational model, and it is not surprising that it does well and is able to work with nonstationarity. 

Across deep learning methods, there is no pattern as to whether the forecast error is better in the stationary or nonstationary case, though there is a pattern within each method. Therefore, it is likely to be best to back test all four configurations for any model before applying it in practice. It is interesting to note that the MAPE decreases as the maturity of the interest rate increases (see the progression of forecast errors for TimeGPT). This is likely because the yield curve is more stable for longer maturities.

\begin{table*}[t]
\centering
\caption{\label{deeplearning_metrics} \small Metrics for the deep learning configurations. Highlighted values are the configurations for each model with the best performance. Bolded is the best of all models.} 
\resizebox{\textwidth}{!}{
\begin{tabular}{l|cc|cc|cc|cc|cc|cc}
\hline & \multicolumn{2}{|c}{ DGS1 } & \multicolumn{2}{c}{ DGS2 } & \multicolumn{2}{c}{ DGS5 } & \multicolumn{2}{c}{ DGS7 } & \multicolumn{2}{c}{ DGS10 }& \multicolumn{2}{c}{ DGS30 }\\
& RMSE & MAPE & RMSE & MAPE &  RMSE& MAPE & RMSE & MAPE& RMSE & MAPE& RMSE & MAPE \\
\hline RNN (sliding stationary) &\underline{\hl{0.146486}}& \underline{\hl{5.691793}}&
\underline{\hl{0.166394}}&5.716609&
\underline{\hl{0.176017}}&4.465607&
\underline{\hl{0.177391}}&\underline{\hl{3.926334}}&
\underline{\hl{0.170646}}&3.410476&
\underline{\hl{0.152589}}&\underline{\hl{2.581476}}\\
RNN (sliding nonstationary) & 0.195552&9.748064&
0.208004&8.089539&
0.212033&5.506938&0.209340&4.693390&0.203377&4.139175&0.189386&3.366322\\
RNN (expanding stationary) & 0.154095&
5.712689&
0.168214&
\underline{\hl{5.490106}}&
0.182967&
\underline{\hl{4.455710}}&
0.181488&
3.894526&
0.176783&
3.447804&
0.161176&
2.670425\\
RNN (expanding nonstationary) & 0.174211&
7.175032&0.186559&6.950657&
0.186624&4.725463&0.183169&
3.983092&0.173481&\underline{\hl{3.400692}}&
0.162370&
2.693776 \\
\hline
NBEATS (sliding stationary) & 0.169360 &6.735220&0.193535&
6.775572&0.212114&5.300780&0.212962&4.618561&0.204240&4.054331&0.183500&3.080803\\
NBEATS (sliding nonstationary) & 0.155120&5.972669&0.172803&\underline{\hl{5.821611}}&\underline{\hl{0.183478}}&\underline{\hl{4.592024}}&\underline{\hl{0.182792}}& \underline{\hl{4.0196230}}&0.178032&\underline{\hl{3.586611}}&0.161656&2.748928 \\
NBEATS (expanding stationary) & 0.173883&
6.486442&0.193031&6.541242&0.208101&5.094602&0.208493&
4.438186&0.200222&3.901296&0.184558&3.030955\\
NBEATS (expanding nonstationary) &\underline{\hl{0.151298}}&\underline{\hl{5.916051}}&\underline{\hl{0.172361}}&
5.843841&0.186442&4.712045&0.186479&4.144330&\underline{\hl{0.177689}}&3.599258&
\underline{\hl{0.160228}}&\underline{\hl{2.739180}} \\
\hline
NHITS (sliding stationary) & 0.196427&8.606049&0.218681&8.168106&0.239196&6.457983&0.236885&5.695617&0.229014&4.949347&0.211299&3.878304\\
 NHITS (sliding nonstationary) &\underline{\hl{0.156482}}&
\underline{\hl{5.983679}}&\underline{\hl{0.173916}}&\underline{\hl{5.860408}}
&\underline{\hl{0.185598}}&\underline{\hl{4.677531}}&\underline{\hl{0.185756}}&
\underline{\hl{4.086771}}&\underline{\hl{0.179829}}&\underline{\hl{3.594320
}}&\underline{\hl{0.160569}}&\underline{\hl{2.717352}}\\
 NHITS (expanding stationary)& 0.176017&
7.021676&0.197292&6.960992&0.209467&5.361165&0.210712&
4.730384&0.205131&4.170214&0.183448&3.091368\\
NHITS (expanding nonstationary) & 0.163171&6.252572&
0.184494&6.245819&0.198605&4.935274&
0.196601&4.267738&0.189513&3.726338&0.169368&2.83406\\

\hline DeepAR (sliding stationary)  & 0.192119&7.581507&0.213584&7.294004&0.233002& 5.806743& 0.236858& 5.025318&0.228642&4.319988&0.213584&3.468202\\
DeepAR (sliding nonstationary) & 0.219236&8.408970&0.240060&7.754574&0.247739&6.125960&0.242207&5.230717&0.235767& 4.593314 &0.209073&3.445667\\
DeepAR (expanding stationary) & \underline{\hl{0.172453}}&
\underline{\hl{7.350338}}&
\underline{\hl{0.194703}}&\underline{\hl{6.805798}}&\underline{\hl{0.208526}}&\underline{\hl{5.551862}}&\underline{\hl{0.206971}}&
\underline{\hl{4.728295}}&\underline{\hl{0.201522}}&
\underline{\hl{4.175998}}&\underline{\hl{0.181019}}&\underline{\hl{3.149804}}\\
DeepAR (expanding nonstationary) &0.212721&
9.348876&0.236940&8.509383&0.252568&6.849793&
0.247249&5.874684&0.239420&5.136509&0.215373&3.853309\\
\hline
Vanilla Transformer (sliding stationary) & 0.168031&
6.658371&0.189859&6.590129&0.201136&5.139067&0.199215&4.424766&0.190739&3.862510&0.173157&2.951177\\
Vanilla Transformer (sliding nonstationary) &0.198712&
7.292155&0.220846&7.033475&0.233620&
5.697577&0.231314&4.953559&0.223127&
4.418951&0.199697&3.378204\\
Vanilla Transformer (expanding stationary) & \underline{\hl{0.158572}}&
7.036351&\underline{\hl{0.179798}}&6.671527&\underline{\hl{0.196566}}&5.304733&\underline{\hl{0.197786}}&
4.651128&\underline{\hl{0.191146}}&4.115878&0.172766&3.083212\\
Vanilla Transformer (expanding nonstationary) & 0.169460&
\underline{\hl{6.423364}}&0.189774&\underline{\hl{6.118753}}&0.199669&
\underline{\hl{4.842441}}&0.202535&\underline{\hl{4.297821}}&0.195714&\underline{\hl{3.790389}}&
\underline{\hl{0.172761}}&\underline{\hl{2.851225}}\\
\hline
Informer (sliding stationary) & 0.161757&6.567544&0.182682&6.604669&0.201996&5.211218&0.202185&4.535645&0.196770&4.010102&
0.177369&3.045935\\

Informer (sliding nonstationary) & 0.165135&
6.411187&0.188657&6.160168&
0.196469&4.841849&0.195355&4.253372&0.189050&3.739071&
0.171406&2.913314\\

Informer (expanding stationary)&\underline{\hl{0.160592}}&
\underline{\hl{6.140962}}&\underline{\hl{0.181482}}&
6.056592&0.195449&4.919052&0.195497&4.337823&0.188660&
3.817651&0.172038&
2.930998\\
Informer (expanding nonstationary) &0.164961&
6.191474&0.182133&
\underline{\hl{5.907332}}&\underline{\hl{0.195373}}&
\underline{\hl{4.705156}}&\underline{\hl{0.192961}}&
\underline{\hl{4.132589}}&\underline{\hl{0.187783}}&
\underline{\hl{3.684220}}&\underline{\hl{0.170620}}&\underline{\hl{2.833431}}\\
\hline
PatchTST (sliding stationary) & 0.160592&6.947785&0.195449&
6.622958&0.195449&5.154602&0.195497&4.484637&0.188660&3.857123&
0.172038&2.962577\\
PatchTST (sliding nonstationary) & 0.154030&5.955978&
0.172140&5.813934&0.185133&
4.668095&\underline{\hl{0.182121}}&4.025542&0.176372&3.556664&\underline{\hl{0.158867}}&2.707418\\
PatchTST (expanding stationary) & 0.168842&6.690186&0.187639&
6.477896&0.205637&5.135185&0.204779&4.446965&0.196693&
3.886160&0.176897&2.933590\\
PatchTST (expanding nonstationary) &\underline{\hl{0.152471}}&\underline{\hl{5.731149}}&\underline{\hl{0.170355}}&\underline{\hl{5.556025}}&\underline{\hl{0.183704}}&\underline{\hl{4.537304}}&0.183254&\underline{\hl{3.961829}}&\underline{\hl{0.175795}}&\underline{\hl{3.452532}}&0.158956&\underline{\hl{2.653176}}\\
\hline
TFT (sliding stationary) &\underline{\hl{0.157429}}&\underline{\hl{5.874464}}&\underline{\hl{0.178096}}
&\underline{\hl{5.962244}}&0.195562&\underline{\hl{4.815003}}&\underline{\hl{0.193904}}&\underline{\hl{4.213565}}&\underline{\hl{0.190366}}&\underline{\hl{3.762933}}&
\underline{\hl{0.171455}}&\underline{\hl{2.875232}}\\
TFT (sliding nonstationary) &0.179963&6.779674&0.198092&6.366203&0.205914&5.093868&
0.204881&4.537172&0.198841&3.952223&0.176219&2.992915\\

TFT (expanding stationary) & 0.159421&6.709400&0.181031& 6.476538&
0.199413& 5.20641& 0.202069& 4.61318 &0.19785& 4.094073&0.174754&3.058329\\
TFT (expanding nonstationary) & 0.160591&   6.947784&0.1814820&6.622957&
\underline{\hl{0.195449}}& 5.154601&0.195496 & 4.484636&
0.188660& 3.857123&0.172038&2.962577\\
\hline

TimeGPT (sliding nonstationary) & \textbf{0.142020}&\textbf{5.289395}& 
\textbf{0.159870}& \textbf{5.192356}&
\textbf{0.171269} & \textbf{4.261029}&
\textbf{0.170635}&\textbf{3.740923}&\textbf{0.164447 }&\textbf{3.292777}& \textbf{0.147716}&  \textbf{2.519419}\\
\hline
\end{tabular}}
\end{table*}

\subsection{Analysis of Classical Machine Learning \& Econometrics}

Table \ref{ml_metrics} shows the results for the traditional methods, classical machine learning methods, and naive benchmarks. 
 Table \ref{tab:var2} in Appendix \ref{std_tables} shows the standard deviation of these results.
Among the traditional methods, ARIMA sliding, ARIMA expanding, and the Naive Forecast have the best performance in the overall average for both the RMSE and MAPE. For ARIMA, by exact value, the expanding ARIMA model does better for some bonds, and sliding ARIMA model does better in some areas. But the difference in MAPE are all less than 0.1$\%$, and many are even less than 0.01$\%$. So in the overall average, there is no clear winner, and they are effectively tied. We can look at the time blocks for ARIMA graphically in the Appendix Figure \ref{fig:arimacomp}. Looking at the RMSE, graphically there does not seem to be too much of a noticeable difference for sliding vs. expanding; the dots representing the RMSE value at the i-th hyperparameter time block nearly completely overlap. The DHR-ARIMA models also come close to the performance of the ARIMA models. The DHR-ARIMA model is a model that combines the ARIMA model with Fourier terms.


For all of the ensemble ML methods (Random Forests, LGBM and XGBoost), the expanding window configurations beat the sliding window ones for both RMSE and MAPE. For Random Forests, however, the MAPE differences are less than 0.1$\%$. For LGBM and XGBoost, the difference in MAPE are all larger than 0.1\%; for LGBM it ranges from 0.28\% to 1.7\% and for XGBoost it ranges from 0.2\% to 1.6\%. So clearly, more data gives better performance for these methods. Overall, the ML methods are unable to outperform traditional econometric methods.

\begin{table*}[t]
\centering
\caption{\label{ml_metrics} \small Metrics for classical machine learning, traditional time series, and benchmark models for both sliding and expanding window configurations. Highlighted values are the configurations for each model with the best performance. } 
\resizebox{\textwidth}{!}{
\begin{tabular}{l|cc|cc|cc|cc|cc|cc}
\hline & \multicolumn{2}{|c}{ DGS1 } & \multicolumn{2}{c}{ DGS2 } & \multicolumn{2}{c}{ DGS5 } & \multicolumn{2}{c}{ DGS7 } & \multicolumn{2}{c}{ DGS10 }& \multicolumn{2}{c}{ DGS30 }\\
& RMSE & MAPE & RMSE & MAPE & RMSE & MAPE & RMSE & MAPE & MAPE & RMSE & MAPE & RMSE \\
\hline Naive & 0.135770&5.055572&0.152728&4.993154&
0.163252&4.049806&0.162623&3.547320&0.156592&3.117768&0.141502&2.395196\\
\hline
Naive Mean(sliding)& \underline{\hl{1.533966}} &\underline{\hl{148.892752}} &\underline{\hl{1.406859}} &\underline{\hl{89.746779}}&\underline{\hl{1.128839} }&\underline{\hl{40.778551}} &
\underline{\hl{1.014991}}&\underline{\hl{29.009124}} &\underline{\hl{0.911509}} &\underline{\hl{0.224073}} &\underline{\hl{0.750931}}&\underline{\hl{14.436432}} \\
Naive Mean (expanding)  &3.503505 &793.977015 &3.504329 &415.631810&
3.380958 & 173.709844 &3.301792&128.761174 &
3.214526 &106.0753242&2.952753&74.544802\\
\hline
Seasonal Naive &0.157473&5.864617&0.177100&5.838530&0.190007&4.734742&
0.188553&4.128603&0.181263&3.627423&0.162442&2.757525 \\
\hline
Random Forests (sliding) & 0.150790&
6.196810&0.171023&5.894464&0.180937&4.599250&0.177776&3.944089&0.172608&3.492104&0.154956&
2.653543\\
Random Forests (expanding) &\underline{\hl{0.146412}}&\underline{\hl{5.74374}}&
\underline{\hl{0.165113}}&\underline{\hl{5.739487}}&\underline{\hl{0.176855}}
&\underline{\hl{4.517249}}&
\underline{\hl{0.176312}}&\underline{\hl{3.935435}}&\underline{\hl{0.168709}}
&\underline{\hl{3.397517}}&\underline{\hl{0.153621}}&\underline{\hl{2.617225}} \\
\hline
XGBoost (sliding) &0.165700&7.250983&0.187430&6.488116&0.201199&
5.154135&0.194842&4.376742&0.190484&3.822953&0.173862&2.967910\\
XGBoost (expanding) &\underline{\hl{0.152398}}&\underline{\hl{5.616206}}&
\underline{\hl{0.175870}}&\underline{\hl{5.809686}}&\underline{\hl{0.186491}}
&\underline{\hl{4.635582}}&\underline{\hl{0.184002}}&\underline{\hl{4.008400}}&
\underline{\hl{0.178698}}&\underline{\hl{3.521038}}&\underline{\hl{0.160632}}&\underline{\hl{2.714995}}\\
\hline
LGBM (sliding) & 0.161913&7.026354&0.177675&6.176718&0.189634&4.799284&
0.185725&4.097728&0.181207&3.652131&0.164472&2.793609 \\
LGBM (expanding) &\underline{\hl{0.142607}}&\underline{\hl{5.256113}}&
\underline{\hl{0.160504}}&\underline{\hl{5.260603}}&\underline{\hl{0.172458}}&
\underline{\hl{4.264067}}&\underline{\hl{0.171610}}&\underline{\hl{3.743909}}&
\underline{\hl{0.165295}}
&\underline{\hl{3.257688}}&\underline{\hl{0.148907}}&\underline{\hl{2.504541}}\\ \hline

ARIMA (sliding) & 0.135890&\underline{\hl{5.055572}}&0.152612&
\underline{\hl{5.005200}}&\underline{\hl{0.163123}}&4.059200&\underline{\hl{0.162586}}&\underline{\hl{3.555000}}&\underline{\hl{0.156626}}&\underline{\hl{3.120700}}&\underline{\hl{0.141562}}&\underline{\hl{2.397700}}\\
ARIMA (expanding)&\underline{\hl{0.135159}}&
5.1487& 0.152409&5.0561&0.163337&4.0585&0.162833&
3.5543&0.156754&3.1232&0.141746&2.4018\\ \hline
DHR-ARIMA (sliding) &0.138657&6.2857&0.156686&\underline{\hl{5.1705}}&
\underline{\hl{0.166255}}&\underline{\hl{4.1621}}&\underline{\hl{0.165264}}&\underline{\hl{3.6306}}&
\underline{\hl{0.158840}}&\underline{\hl{3.1763}}&\underline{\hl{0.143420}}&\underline{\hl{2.4417}}
\\
DHR-ARIMA (expanding) & \underline{\hl{0.137952}}&
\underline{\hl{5.929800}}&\underline{\hl{0.154671}}&5.227900&\underline{\hl{0.166255}}&\underline{\hl{4.1621}}&\underline{\hl{0.165264}}&
\underline{\hl{3.6306}}&\underline{\hl{0.158840}}&\underline{\hl{3.1763}}&\underline{\hl{0.143420}}&\underline{\hl{2.4417}}\\
\hline
diff+VAR (sliding) & 0.138799&
5.670032&0.153870&5.101855&0.163995&4.071155&0.163209&3.565666&0.157270&3.134817&0.142127&2.407170\\
diff+VAR (expanding) &\underline{\hl{0.1368373}}
&\underline{\hl{5.198148}}&\underline{\hl{0.153148}}
&\underline{\hl{5.034782}}&\underline{\hl{0.163663}}
&\underline{\hl{4.053175}}&\underline{\hl{0.162886}}&\underline{\hl{3.547698}}
&\underline{\hl{0.157007}}&\underline{\hl{3.122647}}&\underline{\hl{0.141831}}&\underline{\hl{2.400855}}
\\ \hline
VECM (sliding) & \underline{\hl{0.1533478}}&\underline{\hl{6.446511}}&\underline{\hl{0.16965499}}&\underline{\hl{5.908182}} &0.18156569 &\underline{\hl{4.576077}}&0.18084767&3.984029 
 &0.17377951&3.508151  &0.15539558 &2.666769\\
VECM (expanding) & 0.15623516&7.701282 &0.17347999&6.220672& \underline{\hl{0.1798443}}&4.633831 & \underline{\hl{0.17667}}&\underline{\hl{3.937529}}  &  \underline{\hl{0.16877586}}&\underline{\hl{3.410169}}& \underline{\hl{0.15207266}}&\underline{\hl{2.6136}} \\ \hline
\end{tabular}}
\end{table*}

\subsection{Final Model Performance Table Analysis}

Table \ref{leaderboard} is the final model performance table that compares results from the best configurations.
 Table \ref{tab:var3} in the Appendix \ref{std_tables} shows the standard deviation of these results.
On overall average, ARIMA expanding window, ARIMA sliding window or Naive forecasts do the best, suggesting that for yield curve forecasting, traditional econometric approaches are still the method of choice. For DGS1 and DGS2, ARIMA expanding window does the best, but Naive Forecast does the best on the rest. No other methods has beat the Naive Forecast. The intuition for this can be understood by the fact that at a daily frequency, the yield value typically stays stable from day to day with very mild fluctuations, except when there is some policy change or economic factor causing a sharp jump. So the process is close to a martingale. 

Of the classical ML methods, expanding LGBM does the best. Of the DL methods, TimeGPT does the best, while sliding stationary RNN comes in second. The top three ML algorithms are TimeGPT, expanding LGBM, and the sliding stationary RNN.

\begin{table*}
\centering
\caption{\label{leaderboard} \small The final model performance table. Highlighted values are the models with the best performance.} 
\resizebox{\textwidth}{!}{
\begin{tabular}{l|cc|cc|cc|cc|cc|cc}
\hline & \multicolumn{2}{|c}{ DGS1 } & \multicolumn{2}{c}{ DGS2 } & \multicolumn{2}{c}{ DGS5 } & \multicolumn{2}{c}{ DGS7 } & \multicolumn{2}{c}{ DGS10 }& \multicolumn{2}{c}{ DGS30 }\\
& RMSE & MAPE & RMSE & MAPE & RMSE & MAPE & RMSE & MAPE& RMSE & MAPE & RMSE & MAPE\\
\hline
\hline Naive & 0.13577&$\hl{\textbf{5.055572}}$&0.152728&$\hl{\textbf{4.993154}}$&
0.163252&$\hl{\textbf{4.049806}}$&0.162623&$\hl{\textbf{3.547320}}$&$\hl{\textbf{0.156592}}$&$\hl{\textbf{3.117768}}$&$\hl{\textbf{0.141502}}$&$\hl{\textbf{2.395196}}$\\
\hline
Naive Mean (sliding)& 1.533966 &148.892752 &1.406859 &89.746779&1.128839 &40.778551 &
1.014991 &29.009124 &0.911509 &0.224073 &0.750931&14.436432 \\
\hline
Seasonal Naive &0.157473&5.864617&0.177100&5.838530&0.190007&4.734742&
0.188553&4.128603&0.181263&3.627423&0.162442&2.757525 \\
\hline

ARIMA (sliding) & 0.135890&
$\hl{\textbf{5.055572}}$&0.152612&
5.005200&$\hl{\textbf{0.163123}}$&4.059200&$\hl{\textbf{0.162586}}$&
3.555000&0.156626&3.120700&0.141562&2.397700\\
ARIMA (expanding)&$\hl{\textbf{0.135159}}$&
5.1487&$\hl{\textbf{0.152409}}$&5.0561&0.163337&4.0585&0.162833&
3.5543&0.156754&3.1232&0.141746&2.4018\\ \hline
DHR-ARIMA (sliding) &0.138657&6.2857&0.156686&5.1705&
0.166255&4.1621&0.165264&3.6306&
0.158840&3.1763&0.143420&2.4417
\\
DHR-ARIMA (expanding) & 0.137952&
5.929800&0.154671&5.227900&0.166255&4.162100&0.165264&
3.630600&0.158840&3.176300&0.143420&2.441700\\

\hline
diff+VAR (expanding) &0.1368373 &
5.198148&0.153148&5.034782&0.163663&4.053175&0.162886&3.547698&0.157007&
3.122647&0.141831&2.400855

\\ \hline
VECM (sliding) & 0.1533478 &6.446511 &0.16965499&5.908182 &0.18156569 &4.576077 &0.18084767&3.984029 
 &0.17377951&3.508151  &0.15539558 &2.666769\\
VECM (expanding) & 0.15623516&7.701282 &0.17347999&6.220672& 0.1798443&4.633831 & 0.17667&3.937529  &  0.16877586&3.410169& 0.15207266&2.6136 \\
\hline Random Forests (expanding)  &0.146412&5.74374&0.165113
&5.739487&0.176855&4.517249&
0.176312&3.935435&0.168709&3.397517&0.153621&2.617225 \\

\hline XGBoost (expanding) &0.152398&5.616206&
0.175870&5.809686&0.186491&4.635582&0.184002&4.008400&
0.178698&3.521038&0.160632&2.714995\\

\hline LGBM (expanding) &0.142607&5.256113&
0.160504&5.260603&0.172458&4.264067&0.171610&3.743909&
0.165295&3.257688&0.148907&2.504541\\ \hline

\hline\hline RNN (sliding stationary) &0.146486&5.691793&
0.166394&5.716609&
0.176017&4.465607&
0.177391&3.926334&
0.170646&3.410476&
0.152589&2.581476\\

RNN (expanding stationary) & 0.154095&
5.712689&
0.168214&
5.490106&
0.182967&
4.455710&
0.181488&3.894526&
0.176783&3.447804&0.161176&2.670425\\
RNN (expanding nonstationary) & 0.174211&
7.175032&0.186559&6.950657&0.186624&
4.725463&0.183169&3.983092&0.173481&
3.400692&0.162370&2.693776 \\
\hline
NBEATS (sliding nonstationary) & 0.155120&5.972669&0.172803&5.821611&0.183478&4.592024&0.182792& 4.0196230&0.178032&3.586611&0.161656&2.748928 \\
NBEATS (expanding nonstationary) &0.151298&5.916051&0.172361&
5.843841&0.186442&4.712045&0.186479&4.144330&0.177689&3.599258&
0.160228&2.739180 \\

\hline

 NHITS (sliding nonstationary) &0.156482&
5.983679&0.173916&5.860408&0.185598&4.677531&0.185756&
4.086771&0.179829&3.594320&0.160569&2.717352\\
\hline
DeepAR (expanding stationary)
&0.172453&7.350338&0.194703&6.805798&0.208526&5.551862&0.206971&4.728295&0.201522&
4.175998&
0.181019&
3.149804\\
\hline
Vanilla Transformer (sliding nonstationary) &0.198712&
7.292155&0.220846&7.033475&0.233620&
5.697577&0.231314&4.953559&0.223127&
4.418951&0.199697&3.378204\\
Vanilla Transformer(expanding stationary) &0.158572&
7.036351&0.179798&6.671527&0.196566&5.304733&0.197786&
4.651128&0.191146&4.115878&0.172766&3.083212\\
\hline
Informer(expanding stationary)&0.160592&
6.140962&0.181482&
6.056592&0.195449&4.919052&0.195497&4.337823&0.188660&
3.817651&0.172038&
2.930998\\
Informer (expanding nonstationary) &0.164961&
6.191474&0.182133&
5.907332&0.195373&
4.705156&0.192961&
4.132589&0.187783&
3.684220&0.170620&
2.833431\\
\hline
PatchTST (sliding nonstationary) & 0.154030&5.955978&
0.172140&5.813934&0.185133&
4.668095&0.182121&4.025542&0.176372&3.556664&0.158867&2.707418\\
PatchTST (expanding nonstationary) &0.152471&5.731149&0.170355&5.556025&0.183704&4.537304&0.183254&3.961829&0.175795&3.452532&0.158956&2.653176\\
\hline
TFT (sliding stationary) & 0.157429&5.874464&
0.178096&5.962244&0.195562&4.815003&0.193904&
4.213565&0.190366&
3.762933&0.171455&
2.875232\\
TFT (expanding nonstationary) & 0.160591&   6.947784&
0.1814820&6.622957&
0.195449& 5.154601&
0.195496 & 4.484636&
0.188660& 3.857123&
0.172038&2.962577\\
\hline
TimeGPT (sliding nonstationary) & 0.142020&5.289395& 
0.159870& 5.192356&
0.171269 & 4.261029&
0.170635&3.740923&0.164447 &3.292777& 0.147716&  2.519419\\
\hline
\end{tabular}}
\end{table*}

\begin{table*}[t]
\centering
\caption{\label{tft_patchtst} \small The TFT and PatchTST against Naive and ARIMA on Time Block 8.} 
\resizebox{\textwidth}{!}{
\begin{tabular}{l|cc|cc|cc|cc|cc|cc}
\hline & \multicolumn{2}{|c}{ DGS1 } & \multicolumn{2}{c}{ DGS2 } & \multicolumn{2}{c}{ DGS5 } & \multicolumn{2}{c}{ DGS7 } & \multicolumn{2}{c}{ DGS10 }& \multicolumn{2}{c}{ DGS30 }\\
& RMSE & MAPE & RMSE & MAPE & RMSE & MAPE & RMSE & MAPE& RMSE & MAPE & RMSE & MAPE\\
\hline
\hline Naive &0.125155&8.080886
&0.157298&7.236926
&0.16938&5.734847
&0.164981&5.147716
&0.15713&4.787099
&0.140188&3.833706\\
\hline
ARIMA (sliding)&0.120157&8.48615
&$\mathbf{0.153077}$&$\mathbf{7.109102}$
&0.16938&5.734847
&0.164981&5.147716
&0.157465&4.801335
&0.140546&3.848617\\
ARIMA (expanding)&$\mathbf{0.119377}$&$\mathbf{7.993785}$
&0.154068&7.270121
&0.168742&5.728651
&0.164294&5.13737
&0.15666&4.786389
&0.140324&3.842605\\ \hline
PatchTST (expanding nonstationary) 
&0.123822 &8.056854 &
0.155287&7.251545
&$\mathbf{0.167572}$ &$\mathbf{5.690937}$
& $\mathbf{0.163061}$&$\mathbf{5.081589}$
& $\mathbf{0.155472}$ &$\mathbf{4.737182}$ 
&$\mathbf{0.139728}$ &$\mathbf{3.820803}$\\
\hline
TFT (expanding nonstationary) 
& 0.120108  &8.017696  
 & 0.155279 & 7.282278 
& 0.169733 &5.784258 
&  0.165366 &5.183267 
& 0.157947 & 4.832981 
& 0.140756 & 3.846431  
\\
\hline
\end{tabular}
}
\end{table*} 

However, if we observe the different hyperparameter time blocks, we notice that in some cases, PatchTST and TFT can do better, especially with nonstationary expanding data. Table \ref{tft_patchtst} shows the TFT and PatchTST (both expanding nonstationary) against the Naive Forecast and ARIMA on Time Block 8.  For DGS1, nonstationary expanding TFT beats the Naive Forecast and sliding ARIMA, but not  expanding ARIMA. However, these are very  close scores. In Time Block 8 for DGS5, DG7, DGS10, and DGS30, the nonstationary expanding PatchTST outperforms the other methods; however, these are very small differences, such as 0.03\% or 0.05\%. This could be due to more data or could be due to the time block. Therefore, we can observe its performance over time to see if it improves over time relative to the Naive Forecast and ARIMA; and we can compare it to the sliding window methods (see Figure \ref{fig:patchtstbest} for a graph). The PatchTST sliding window also performs well in this time block, but is worse than the expanding nonstationary. So some of the effects can be due to both the type of random process the last time block is, along with having more data from being an expanding window configuration.



Figure \ref{fig:arimacomp} shows the performance of the ARIMA model over time. Figures  \ref{fig:patchtstbest}, \ref{fig:TFTbest}, and \ref{fig:rnn_best} in Appendix \ref{rmse_mape_plots} compare the performance of the best configurations over time of some of the deep learning models vs ARIMA and the Naive Forecast. The upper half of the figures shows the evolution of RMSE over time and the bottom half shows the evolution of MAPE over time. The RMSE falls over time and the MAPE seems to increase over time. This is simply because interest rates have fallen to lower levels on average in the sample period, as can be seen in Figure \ref{fig:dgs10}. Hence, error levels have fallen but percentage error increases. 

We have also undertaken similar analyses on the overall averages for the forecast horizons of 20 and 60 business days. Overall, it showed that the Naive Forecast usually outperforms all other methods, with ARIMA close behind. This affirms that traditional econometric methods are still the best for this problem. The pre-trained models like TimeGPT were only used in zero-shot mode without fine-tuning and may not be competitive with econometric models like ARIMA that are fitted each time a forecast is made. The deep learning models we tried may be underperforming on account of overfitting the data leading to poor out of sample performance.

\section{Conclusion}

The paper addresses a major artifact in finance, the yield curve, that underpins all trading in bonds. Bond markets are much larger than equity markets as noted in Section \ref{sec:data} and thus, the analyses in this paper will speak to a large audience of financial market participants. Yield curve slopes are empirically known harbingers of recessions, i.e.,  when the yield curve slopes downwards, it is predictive of a recession about 18-24 months out, with its attendant financial and economic consequences for all citizens. For this, the ability to forecast the slope from the individual yield forecasts is provided in the methodology. The apt choice of forecasts will also be of interest to regulators, especially the Federal Reserve and the Treasury Department, who manage yields and are impacted by interest rates.

In this paper we had four major goals:
(1) To compare the performance of  machine learning (classical ML  and deep learning) vs. traditional econometrics/time-series methods on time series of Treasury yields. We used RMSE and MAPE to evaluate the performance. 
(2) Check the performance of models on multiple forecast horizons and over different rolling time periods.
(3) Determine whether expanding vs. sliding window training works better.
(4) Compare stationary vs. nonstationary inputs specifically for deep learning algorithms on this dataset.
 

In Section \ref{results}, we implemented experiments to  comparatively assess these goals over a 47 year daily time series. Our analyses show that a traditional econometrics algorithm (ARIMA) and a naive benchmark model performed the best on average. Therefore, deep learning approaches failed to outperform traditional (and arguably simpler) econometric methods. Deep learning methods outperformed other methods only in the last time block, in that PatchTST performed better with expanding training windows, albeit by a small margin. Classical machine learning models also failed to beat traditional models. In the cases where we forecasted 20 and 60 days ahead, these broad results remained unchanged.

For our goal of exploring whether deep learning model perform better under stationarity versus non-stationarity data, the RNN model performed better with a stationary sliding configuration; PatchTST and TFT performed better with a nonstationary one.

\subsection{Extensions and Areas of Improvement}

There are many ways in which the analyses in this paper may be extended. {\it First}, we can use more exogenous variables as features. For example, sentiment analysis of social media and news may be used to aid yield curve forecasting for recession prediction. Additionally, incorporating exogenous variables such as macroeconomic indicators, market sentiment, and other relevant financial metrics could provide more context and improve the accuracy of the forecasts.
{\it Second}, probabilistic forecasting methods, such as conformal prediction \cite{manokhin2023conformal}, can be applied to quantify uncertainty in our forecasts. Since we are comparing several methods, we are using point forecasting instead of using prediction intervals/probabilistic forecasting, because all methods do not support easy generation of prediction error intervals.
{\it Third}, it is possible to use null hypothesis significance testing to compare results. This introduces the potential complication of dealing with multiple comparisons.
{\it Fourth}, future work could also explore the use of forecast combination methods that combine the strengths of different models to achieve better performance.
{\it Fifth}, we may use more recent foundation models for forecasting that are now being trialed. For TimeGPT, we did not explore stationary vs. nonstationary inputs; we could also explore how fine tuning can change the results.
{\it Sixth}, we can use different data frequencies, in particular weekly data. This may improve forecasts if the daily data is more noisy than weekly data. 

Another extension we will explore is using multivariate estimation. VAR and VECM were tested, and those are proper multivariate models. However, regular ARIMA has to be used in a univariate manner, and many of the deep learning models used were implemented by {\tt Nixtla} as ``global univariate models,'' where a single model is  trained on all the series. However, for fairness of comparison across all models, we did not combine all the time series in a multivariate analysis. It is possible therefore to expect some improvement when this augmentation is done. Additional results not included in this paper are available in \citet{AmanThesis2024}. Another multifactor approach that evidences good results is the work of \citet{ang_no-arbitrage_2003}.

In summary, this paper compared machine learning models with traditional econometrics models and naive benchmarks. The analysis finds that  ARIMA and the Naive method come out on top. Amongst the machine learning models, TimeGPT and expanding LGBM performed the best. We have also compared the sliding window approach with the expanding window approach, along with nonstationary input vs. stationary inputs for deep learning; the results for these vary depending on the model. Overall, yield curves may still be best forecast using traditional econometric methods, which are simple and inexpensive to apply. 



\bibliographystyle{plainnat} 
\bibliography{ts}

@article{ang_no-arbitrage_2003,
	title = {A no-arbitrage vector autoregression of term structure dynamics with macroeconomic and latent variables},
	volume = {50},
	issn = {0304-3932},
	url = {https://www.sciencedirect.com/science/article/pii/S0304393203000321},
	doi = {10.1016/S0304-3932(03)00032-1},
	number = {4},
	urldate = {2025-07-16},
	journal = {Journal of Monetary Economics},
	author = {Ang, Andrew and Piazzesi, Monika},
	month = may,
	year = {2003},
	pages = {745--787},
}

@article{Vaswani2017,
  author    = {Vaswani, Ashish and Shazeer, Noam and Parmar, Niki and Uszkoreit, Jakob and Jones, Llion and Gomez, Aidan N. and Kaiser, Lukasz and Polosukhin, Illia},
  title     = {Attention Is All You Need},
  journal   = {arXiv},
  year      = {2017},
  month     = {December},
  volume    = {1706.03762},
  url       = {http://arxiv.org/abs/1706.03762}
}

@article{Singh2022,
  author    = {A. Singh and T. Ogunfunmi},
  title     = {An Overview of Variational Autoencoders for Source Separation, Finance, and Bio-Signal Applications},
  journal   = {Entropy},
  year      = {2022},
  volume    = {24},
  number    = {1},
  pages     = {55},
  url       = {https://doi.org/10.3390/e24010055}
}

@misc{Bommasani2021,
  author    = {R. Bommasani and others},
  title     = {On the Opportunities and Risks of Foundation Models},
  year      = {2021},
  howpublished       = {\url{https://arxiv.org/abs/2108.07258}},
  note      = {Accessed: May 24, 2024}
}

@article{makridakis2020m4,
  author    = {S. Makridakis and E. Spiliotis and V. Assimakopoulos},
  title     = {The M4 Competition: 100,000 time-series and 61 Forecasting Methods},
  journal   = {International Journal of Forecasting},
  year      = {2020},
  volume    = {36},
  number    = {1},
  pages     = {54--74},
  url       = {https://doi.org/10.1016/j.ijforecast.2019.09.004}
}

@article{smyl2020,
  author    = {S. Smyl},
  title     = {A Hybrid Method of Exponential Smoothing and Recurrent Neural Networks for time-series Forecasting},
  journal   = {International Journal of Forecasting},
  year      = {2020},
  volume    = {36},
  number    = {1},
  pages     = {75-85},
  url       = {https://doi.org/10.1016/j.ijforecast.2019.03.017}
}

@article{lim2021,
  author    = {B. Lim and S. Arık and N. Loeff and T. Pfister},
  title     = {Temporal Fusion Transformers for interpretable multi-horizon time-series forecasting},
  journal   = {International Journal of Forecasting},
  year      = {2021},
  volume    = {37},
  number    = {4},
  pages     = {1748-1764},
  url       = {https://doi.org/10.1016/j.ijforecast.2021.03.012}
}

@inproceedings{fedformer,
  author    = {T. Zhou and Z. Ma and Q. Wen and X. Wang and L. Sun and R. Jin},
  title     = {FEDformer: Frequency Enhanced Decomposed Transformer for Long-term Series Forecasting},
  booktitle = {Proceedings of the 39th International Conference on Machine Learning (ICML 2022)},
  year      = {2022},
  volume    = {162},
  pages     = {27268--27286}
}

@inproceedings{autoformer,
  author    = {H. Wu and J. Xu and J. Wang and M. Long},
  title     = {Autoformer: Decomposition Transformers with Auto-Correlation for Long-Term Series Forecasting},
  booktitle = {Advances in Neural Information Processing Systems 34 (NeurIPS 2021)},
  year      = {2021}
}

@article{zhou2021,
  author    = {H. Zhou and others},
  title     = {Informer: Beyond Efficient Transformer for Long Sequence Time-Series Forecasting},
  journal   = {Proceedings of the AAAI Conference on Artificial Intelligence},
  year      = {2021},
  volume    = {35},
  number    = {12},
  pages     = {11106--11115},
  doi       = {10.1609/aaai.v35i12.17325}
}

@inproceedings{nie2023a,
  author    = {Y. Nie and N. H. Nguyen and P. Sinthong and J. Kalagnanam},
  title     = {A time-series is Worth 64 Words: Long-term Forecasting with Transformers},
  booktitle = {The Eleventh International Conference on Learning Representations},
  year      = {2023},
  url       = {https://openreview.net/forum?id=Jbdc0vTOcol}
}

@inproceedings{Liu2022Pyraformer,
  author    = {S. Liu and H. Yu and C. Liao and J. Li and W. Lin and A. X. Liu and S. Dustdar},
  title     = {Pyraformer: Low-Complexity Pyramidal Attention for Long-Range time-series Modeling and Forecasting},
  booktitle = {Proceedings of the International Conference on Learning Representations},
  year      = {2022},
  url       = {https://api.semanticscholar.org/CorpusID:251649164}
}

@article{liu2024itransformerinvertedtransformerseffective,
  author    = {Y. Liu and T. Hu and H. Zhang and H. Wu and S. Wang and L. Ma and M. Long},
  title     = {iTransformer: Inverted Transformers Are Effective for time-series Forecasting},
  journal   = {arXiv preprint},
  year      = {2024},
  eprint    = {2310.06625},
  archivePrefix = {arXiv},
  url       = {https://arxiv.org/abs/2310.06625}
}

@inproceedings{Zeng2023,
  author    = {A. Zeng and M. Chen and L. Zhang and Q. Xu},
  title     = {Are Transformers Effective For time-series Forecasting?},
  booktitle = {Proceedings of the Thirty-Seventh AAAI Conference on Artificial Intelligence (AAAI-23)},
  year      = {2023},
  pages     = {26317},
  isbn      = {978-1-57735-880-0},
  doi       = {10.1609/aaai.v37i9.26317}
}

@article{BlogPost2023,
  author    = {Simhayev, Eli and Rasul, Kashiv and Rogge, Niels},
  title     = {Yes, Transformers Are Effective for Time Series Forecasting (+ Autoformer)},
  journal   = {Hugging Face Blog},
  year      = {2023},
  month     = {June},
  day       = {16},
  url       = {https://huggingface.co/blog/autoformer}
}

@misc{Garza2024,
  author    = {A. Garza and C. Challu and M. Mergenthaler-Canseco},
  title     = {TimeGPT-1},
  year      = {2024},
  eprint    = {2310.03589},
  archivePrefix = {arXiv},
  howpublished       = {\url{https://arxiv.org/abs/2310.03589}}
}

@article{Sen2024,
  author    = {R. Sen and Y. Zhou},
  title     = {A Decoder-Only Foundation Model for Time-Series Forecasting},
  journal   = {Google Research Blog},
  year      = {2024},
  month     = {February},
  day       = {2},
  url       = {https://research.google/blog/a-decoder-only-foundation-model-for-\\time-series-forecasting/}
}

@article{Woo2024,
  author    = {G. Woo and C. Liu and D. Sahoo and C. Xiong},
  title     = {Moirai: A Time-Series Foundation Model for Universal Forecasting},
  journal   = {Salesforce AI Research Blog},
  year      = {2024},
  month     = {January},
  day       = {24},
  url       = {https://blog.salesforceairesearch.com/moirai/}
}

@misc{Ansari2024,
  author    = {A. F. Ansari and L. Stella and C. Turkmen and X. Zhang and P. Mercado and H. Shen and O. Shchur and S. S. Rangapuram and S. P. Arango and S. Kapoor and J. Zschiegner and D. C. Maddix and H. Wang and M. W. Mahoney and K. Torkkola and A. G. Wilson and M. Bohlke-Schneider and Y. Wang},
  title     = {Chronos: Learning the Language of Time Series},
  year      = {2024},
  eprint    = {2403.07815},
  archivePrefix = {arXiv},
  howpublished = {\url{https://arxiv.org/abs/2403.07815}}
}

@misc{Jin2024,
  author    = {M. Jin and S. Wang and L. Ma and Z. Chu and J. Y. Zhang and X. Shi and P.-Y. Chen and Y. Liang and Y.-F. Li and S. Pan and Q. Wen},
  title     = {Time-LLM: Time-Series Forecasting by Reprogramming Large Language Models},
  year      = {2024},
  eprint    = {2310.01728},
  archivePrefix = {arXiv},
  howpublished       = {\url{https://arxiv.org/abs/2310.01728}}
}

@article{NelsonSiegel1987,
  author    = {C. R. Nelson and A. F. Siegel},
  title     = {Parsimonious Modeling of Yield Curves},
  journal   = {The Journal of Business},
  year      = {1987},
  volume    = {60},
  number    = {4},
  pages     = {473},
  month     = {January},
  doi       = {10.1086/296409}
}

@book{HyndmanAthanasopoulos2021,
  author    = {R. J. Hyndman and G. Athanasopoulos},
  title     = {Forecasting: Principles and Practice},
  publisher = {Otexts},
  year      = {2021},
  address   = {Lexington, KY},
  url       = {https://otexts.com/fpp3/}
}

@article{HewamalageAckermannBergmeir2022,
  author    = {H. Hewamalage and K. Ackermann and C. Bergmeir},
  title     = {Forecast Evaluation for Data Scientists: Common Pitfalls and Best Practices},
  journal   = {Data Mining and Knowledge Discovery},
  volume    = {37},
  number    = {2},
  pages     = {788--832},
  year      = {2022},
  url       = {https://doi.org/10.1007/s10618-022-00894-5}
}

@misc{MathWorksTrendStationary,
  author       = {MathWorks},
  title        = {Trend-Stationary vs. Difference-Stationary Processes},
  year         = {2024},
  url          = {https://www.mathworks.com/help/econ/trend-stationary-vs-difference-stationary.html},
  note         = {Accessed: Jan. 11, 2024}
}

@misc{MathWorksUnitRoot,
  author       = {MathWorks},
  title        = {Unit Root Nonstationarity},
  year         = {2024},
  url          = {https://www.mathworks.com/help/econ/unit-root-nonstationarity.html},
  note         = {Accessed: Jan. 11, 2024}
}

@book{Hsu2014,
  author       = {H. P. Hsu},
  title        = {Schaum’s Outline Probability, Random Variables, and Random Processes},
  publisher    = {McGraw-Hill Education},
  year         = {2014},
  address      = {New York},
  chapter      = {Random Processes},
  pages        = {200--211}
}

@book{Tsay2010,
  author       = {R. S. Tsay},
  title        = {Analysis of Financial Time Series},
  publisher    = {Wiley},
  year         = {2010},
  address      = {Hoboken, NJ}
}

@misc{Stensby2014,
  author       = {J. Stensby},
  title        = {Chapter 6 - Random Processes},
  institution  = {EE 385, University of Alabama Huntington},
  year         = {2014},
  month        = {November},
  url          = {http://www.ece.uah.edu/courses/ee385/500ch6.pdf},
  note         = {Accessed: March 17, 2023. Alternate link: \url{https://drive.google.com/file/d/1X-2hRmXtF9D-hXuBf1hHs73AmLbNjaj7/view?usp=sharing}}
}

@misc{Chen2022,
  author       = {P. Chen},
  title        = {Vector Error Correction Models with Stationary and Nonstationary Variables},
  year         = {2022},
  month        = {September 14},
  howpublished = {\url{https://ssrn.com/abstract=4218834}},
  note         = {Available at SSRN: \url{http://dx.doi.org/10.2139/ssrn.4218834}}
}

@article{Salinas2020,
  author    = {D. Salinas and V. Flunkert and J. Gasthaus and T. Januschowski},
  title     = {DeepAR: Probabilistic Forecasting with Autoregressive Recurrent Networks},
  journal   = {International Journal of Forecasting},
  volume    = {36},
  number    = {3},
  pages     = {1181--1191},
  year      = {2020},
  doi       = {10.1016/j.ijforecast.2019.07.001},
  url       = {https://doi.org/10.1016/j.ijforecast.2019.07.001}
}

@article{Oreshkin2019,
  author    = {B. N. Oreshkin and D. Carpov and N. Chapados and Y. Bengio},
  title     = {N-BEATS: Neural Basis Expansion Analysis For Interpretable Time Series Forecasting},
  journal   = {arXiv:1905.10437 [cs.LG]},
  year      = {2019},
  url       = {https://doi.org/10.48550/arXiv.1905.10437}
}

@misc{Challu2023,
  author    = {C. Challu and others},
  title     = {NHITS: Neural Hierarchical Interpolation for Time Series Forecasting},
  journal   = {Proceedings of the AAAI Conference on Artificial Intelligence},
  volume    = {37},
  number    = {6},
  pages     = {6989--6997},
  year      = {2023},
  doi       = {10.1609/aaai.v37i6.25854},
  howpublished       = {\url{https://doi.org/10.1609/aaai.v37i6.25854}}
}

@inbook{StockWatson2015,
  author       = {J. H. Stock and M. W. Watson},
  title        = {Introduction to Time Series Regression and Forecasting},
  booktitle    = {Introduction to Econometrics},
  publisher    = {Pearson},
  year         = {2015},
  address      = {Harlow},
  pages        = {568--621}
}

@article{KingmaWelling2014,
  author       = {D. P. Kingma and M. Welling},
  title        = {Auto-Encoding Variational Bayes},
  journal      = {arXiv},
  year         = {2014},
  eprint       = {1312.6114},
  archivePrefix = {arXiv}
}

@inproceedings{Goodfellow2014GAN,
  author       = {I. J. Goodfellow and J. Pouget-Abadie and M. Mirza and B. Xu and D. Warde-Farley and S. Ozair and A. C. Courville and Y. Bengio},
  title        = {Generative Adversarial Nets},
  booktitle    = {Proceedings of the Annual Conference on Neural Information Processing Systems (NIPS)},
  year         = {2014},
  address      = {Montreal, QC, Canada},
  month        = {December},
  pages        = {2672--2680}
}

@inbook{manokhin2023conformal,
  author       = {V. Manokhin},
  title        = {Conformal Prediction for Time Series and Forecasting},
  booktitle    = {Practical Guide to Applied Conformal Prediction in Python},
  publisher    = {Packt},
  year         = {2023},
  month        = {December}
}

@book{ShumwayStoffer2006,
  author    = {R. H. Shumway and D. S. Stoffer},
  title     = {Time Series Analysis and Its Applications},
  publisher = {Springer},
  year      = {2006},
  address   = {New York, NY}
}

@misc{caldeira2013,
  author    = {J. Caldeira and G. V. Moura and A. A. Santos},
  title     = {Predicting the Yield Curve using Forecast Combinations},
  journal   = {SSRN Electronic Journal},
  year      = {2013},
  doi       = {10.2139/ssrn.2311733},
  howpublished       = {\url{https://dx.doi.org/10.2139/ssrn.2311733}}
}

@article{moench2006,
  author    = {E. Moench},
  title     = {Forecasting the Yield Curve in a Data-Rich Environment: A No-Arbitrage Factor-Augmented VAR Approach},
  journal   = {SSRN Electronic Journal},
  year      = {2006},
  doi       = {10.2139/ssrn.676909},
  url       = {https://doi.org/10.2139/ssrn.676909}
}

@article{SwansonXiong2018,
  author    = {N. R. Swanson and W. Xiong},
  title     = {Big Data Analytics in Economics: What Have We Learned So Far, and Where Should We Go From Here?},
  journal   = {Canadian Journal of Economics/Revue canadienne d'économique},
  year      = {2018},
  volume    = {51},
  number    = {3},
  pages     = {695--746}
}

@inproceedings{sambasivan2017,
  author    = {R. Sambasivan and S. Das},
  title     = {A Statistical Machine Learning Approach to Yield Curve Forecasting},
  booktitle = {2017 International Conference on Computational Intelligence in Data Science (ICCIDS)},
  year      = {2017},
  doi       = {10.1109/iccids.2017.8272667}
}

@mastersthesis{reinicke2019,
  author    = {S. Reinicke},
  title     = {Modeling and Forecasting Yield Curves: A Comparison of Published Methods},
  school    = {Ludwig-Maximilians-Universität München},
  year      = {2019},
  type      = {Master's paper},
  address   = {Munich, Germany},
  month     = {April},
  day       = {16}
}

@misc{Zhang2022,
  author    = {W. Zhang and Q. Yang and R. Tian and T. Ye and W. Yao and L. Zhang},
  title     = {Treasury Bond Price and Yield Curve Prediction via No Arbitrage Arguments and Machine Learning},
  year      = {2022},
  month     = {February},
  day       = {2},
  howpublished       = {\url{https://ssrn.com/abstract=4024209}},
  doi       = {10.2139/ssrn.4024209}
}

@mastersthesis{Rahimi2020,
  author    = {K. Rahimi},
  title     = {Forecast Comparison of Financial Models},
  school    = {University of Surrey},
  year      = {2020},
  type      = {PhD thesis}
}

@mastersthesis{oosterlaken2020,
  author    = {J. Oosterlaken},
  title     = {Predicting the US Treasury Yields using Machine Learning Techniques},
  school    = {Erasmus University Rotterdam, Erasmus School of Economics},
  year      = {2020},
  type      = {Master's paper},
  month     = {July},
  day       = {4}
}

@article{Bianchi2020,
  author    = {D. Bianchi and M. Büchner and A. Tamoni},
  title     = {Bond Risk Premiums with Machine Learning},
  journal   = {The Review of Financial Studies},
  year      = {2020},
  volume    = {34},
  number    = {2},
  pages     = {1046--1089},
  doi       = {10.1093/rfs/hhaa062}
}

@mastersthesis{Hoogteijling2020,
  author    = {Tobias Hoogteijling},
  title     = {Forecasting Bond Risk Premia with Machine Learning},
  journal   = {Master's Thesis, Erasmus University, Rotterdam},
  school    = {Erasmus University},
  year      = {2020},
  type      = {Master's paper},
  address   = {Rotterdam, Netherlands; \url{https://thesis.eur.nl/pub/55747/Final_version_thesis.pdf}},
  month     = {October},
  day       = {27}, 
  url = {https://thesis.eur.nl/pub/55747/Final_version_thesis.pdf}
}

@article{Chong2021,
  author    = {W. Chong Pooi Mun and V. Soong},
  title     = {Forecasting Yield Curve with Macro-Driven Models: A Comparison Between Machine Learning and Traditional Statistical Approaches},
  journal   = {RPubs},
  year      = {2021},
  month     = {July},
  url       = {https://rpubs.com/WendyChongPooiMun/YieldCurve#:~:text=For%20long%20term%20forecasting%2C%20the,ideal%20for%20short%20term%20forecasting}
}

@book{LopezDePrado2018,
  author    = {López de Prado, Marcos},
  title     = {Advances in Financial Machine Learning},
  publisher = {Wiley},
  year      = {2018},
  month     = {January},
  isbn      = {978-1119482086},
  url       = {https://www.wiley.com/en-us/Advances+in+Financial+Machine+Learning-p-9781119482086}
}

@mastersthesis{AmanThesis2024,
  author    = {Singh, Aman},
  title     = {Recursive Multistep Forecasting of U.S. Treasury Yield Curves Using Machine Learning and Econometrics},
  school    = {Santa Clara University},
  year      = {2024},
  type      = {Master's Thesis},
  address   = {Santa Clara,USA},
  month     = {September},
  day       = {9}
}

\clearpage
\onecolumn

\appendix

\section{Appendix}

\subsection{Stationarity}
\label{stationarity}

\subsubsection{Weak Stationarity}

A random process $x[n]$ is weakly stationary if it satisfies the following conditions \cite{Hsu2014,Tsay2010}:

\begin{enumerate}
    \item \( \mu_x(n) = \mathbb{E}[x[n]] = \mu \) (constant mean)
    \item \( \gamma_x(k, s) = \gamma_x(|s-k|) \) (autocovariance depends only on the time difference)
    \item Finite variance
\end{enumerate}

The property \( \gamma_x(k, s) = \gamma_x(|s-k|) \) indicates that the autocovariance between the random process at times \( s \) and \( k \) depends only on the time difference \( |s-k| \).


Property 2 implies that the variance is constant.
We will just refer to weakly stationary as stationary.

\subsubsection{Causes of Nonstationarity}

There are multiple causes of nonstationarity 
\cite{MathWorksUnitRoot,MathWorksTrendStationary,HewamalageAckermannBergmeir2022,HyndmanAthanasopoulos2021,Stensby2014,StockWatson2015}.


\begin{enumerate}
    \item  Trends: ``a long-term increase or decrease in the data'' \cite{HyndmanAthanasopoulos2021}.
    \subitem 
    a) Deterministic Trends: A trend that is deterministic. 
    Shocks will have transitory effects.
    \subitem b) Stochastic Trends: the trend is stochastic. Shocks have permanent effects.  Also known as unit roots.
    \item  Structural Breaks:
Abrupt or gradual changes in the population regression coefficients \cite{StockWatson2015}.
    \item Heteroscedasticity: The variance of the series changes.
   \item  Seasonality: This is periodic fluctuations. 
\end{enumerate} 

For stochastic trends, a differencing operation is used to transform the data into having stationarity. For deterministic trends, the trend can be modelled with a regression model and subtracted from the time series. For seasonality, there is seasonal differencing or using seasonal terms.
In this paper, we used the differencing operation to get stationarity in our data.

\subsection{Data Partitioning}
\label{data_partitioning}

One way to evaluate a forecasting algorithm is called the fixed origin method. The data is split into training and testing; after being trained, the algorithm forecasts the next $H$ steps from the testing set; we measure the error on each forecast by a chosen metric.

However, this might not be a rigorous evaluation; it is better to use multiple windows, where the forecast origin moves; this is known as the rolling origin method or the walk-forward method. Sliding window CV and expanding window CV are specific types of the walk-forward method.

\begin{figure}[H]
\centering
({\bf a})\\
\includegraphics[width=3in]{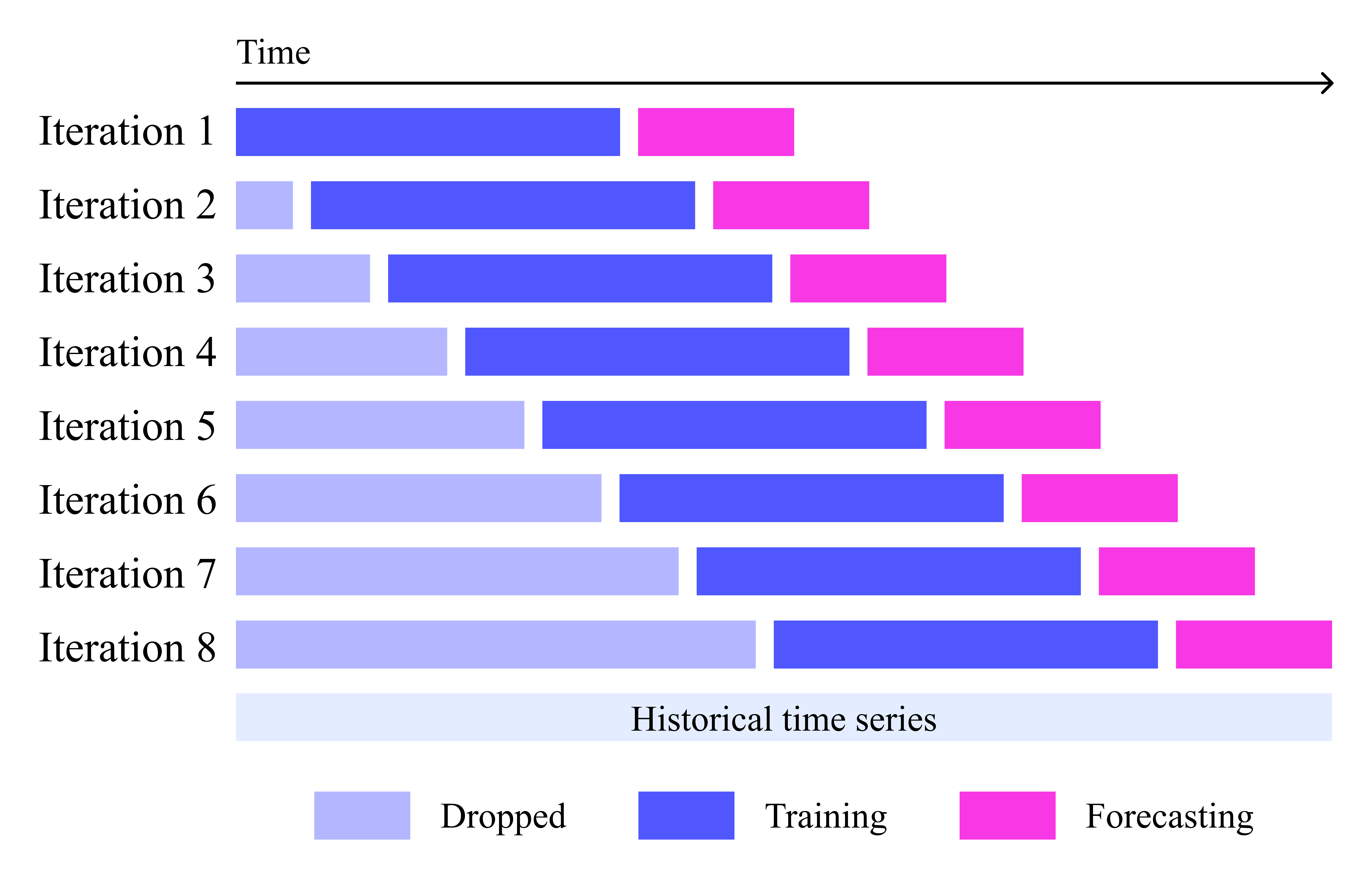}\\
({\bf b})\\
\includegraphics[width=3in]{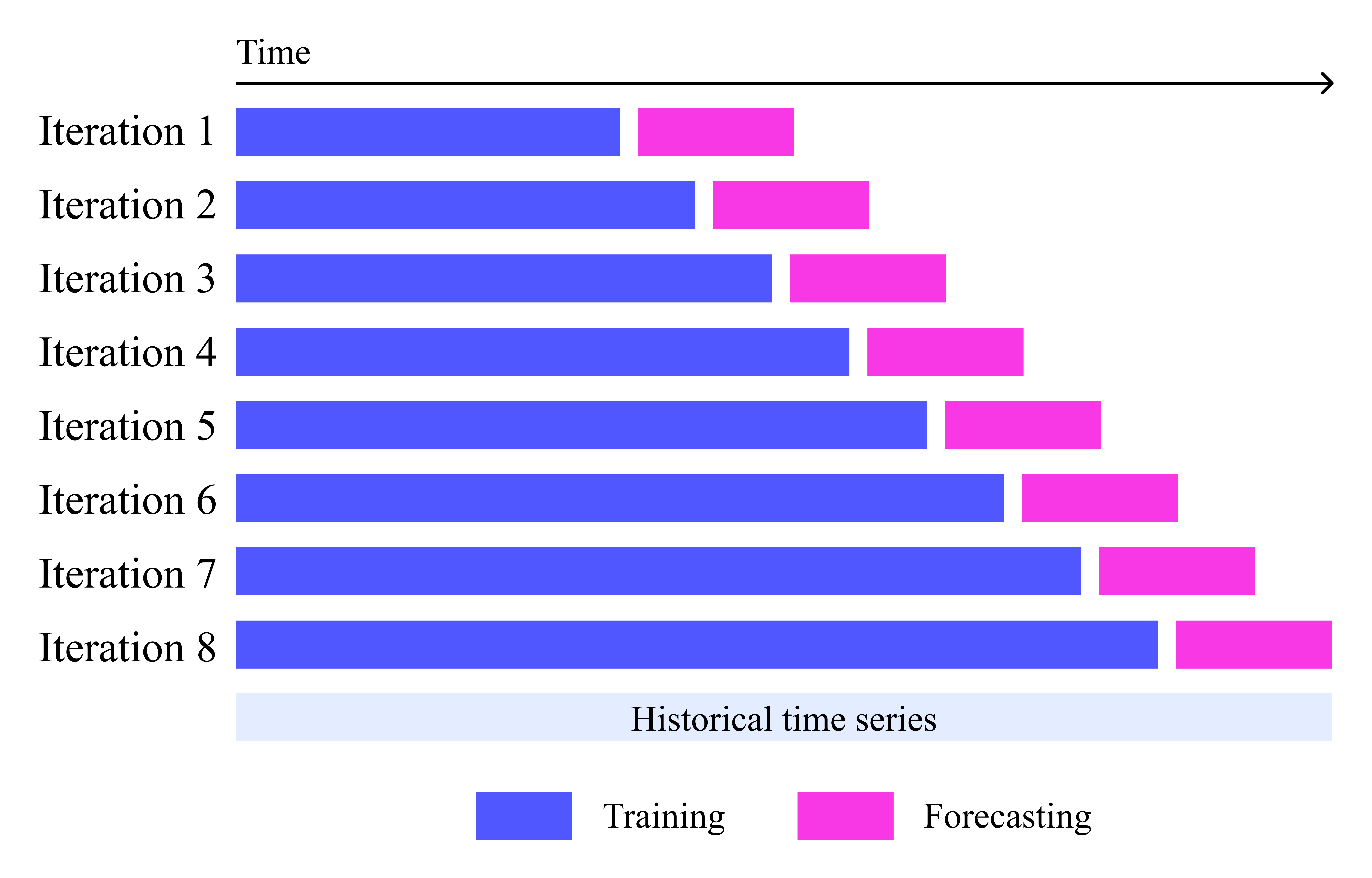}\\
\caption{Sliding window and expanding window (assuming forecast horizon is greater than 1) (\textbf{a}) Sliding window (\textbf{b}) Expanding window. }\label{fig:window}
\end{figure}

In expanding window CV, we start with an initial window of data with size $w$. The model is trained on this window and then forecasts $H$ points, which can be evaluated. Then it expands to include step size $s$, with $w$+$s$ points.  So if the initial window endpoint locations were [0,$w$], then the new window is [0,$w+s$]. It then trains on these points and forecasts $H$ points again.  This is repeated until the forecast horizon reaches the end of the dataset. See Figure \ref{fig:window}.

In sliding window CV, we start with an initial window of data with size $w$, but in this case the size does not change. The model is trained on this window and then forecasts $H$ points, which can be evaluated. Then window moves forward $s$ steps;  So if the initial window endpoint locations were [0,$w$], then the new window is [$s$,$w+s$]. It then trains on these points and forecasts h points again.  This is repeated until the forecast horizon reaches the end of the dataset.

The walk-forward method can be defined as ``a historical simulation of how our algorithm performs in the past'' \cite{LopezDePrado2018}. The advantages of the walk-forward method are the following:
\begin{enumerate}\itemsep0em 
\item There is a historical interpretation.
\item Removes data leakage.
\end{enumerate}

Two major disadvantages are:
\begin{enumerate}\itemsep0em 
\item Going back to the idea of a random  process, the historical path is just one realization of this process. Since the historical path is just one scenario, testing against just one past scenario risks overfitting. 
\item The performance on the historical path by the walk-forward method might not represent the future performance of your algorithm; the algorithm might overfit to the particular order of the  sequence in the sample path.
\end{enumerate}

\subsection{Standard Deviation of Forecast Error Metrics}
\label{std_tables}

In this subsection, we present the standard deviation of the RMSE and MAPE of the forecasts. Table  \ref{tab:var1} contains the standard deviation of the forecast error scores from Table \ref{deeplearning_metrics}, which were the analysis of deep learning methods. we can see that TimeGPT has the lowest standard deviation. Within each model, the best mean of the scores  matches the best standard deviations most of the time, but not in every case.

\begin{table}[H]
\centering
\caption{\label{tab:var1} \small Standard deviation of metrics for the deep learning configurations.} 
\resizebox{\textwidth}{!}{
\begin{tabular}{l|cc|cc|cc|cc|cc|cc}
\hline & \multicolumn{2}{|c}{ DGS1 } & \multicolumn{2}{c}{ DGS2 } & \multicolumn{2}{c}{ DGS5 } & \multicolumn{2}{c}{ DGS7 } & \multicolumn{2}{c}{ DGS10 }& \multicolumn{2}{c}{ DGS30 }\\
& RMSE & MAPE & RMSE & MAPE &  RMSE& MAPE & RMSE & MAPE& RMSE & MAPE& RMSE & MAPE \\
\hline
RNN (sliding stationary) &\underline{\hl{0.194001}}&\underline{\hl{9.371014}}&
\underline{\hl{0.162663}}&7.487174&
\underline{\hl{0.138588}}&4.900849&
\underline{\hl{0.137446}}&3.878843&
\underline{\hl{0.128832}}&\underline{\hl{3.270737}}&
\underline{\hl{0.110987}}&\underline{\hl{2.214017}}
\\
RNN (sliding nonstationary) & 0.210150&23.148221&
0.197456&14.242268&
0.189039&6.998076&
0.184905&5.731761&
0.182635&4.861462&
0.177668&4.364616\\

RNN (expanding stationary) & 0.199035&8.926611&
0.165057&\underline{\hl{6.676691}}&
0.147118&\underline{\hl{4.587371}}&
0.140713&\underline{\hl{3.669374}}&
0.139866&3.306758&
0.126082&2.295027
\\
RNN (expanding nonstationary) & 0.213879&11.242475&
0.184713&24.053279&
0.165088&10.626178&
0.156569&5.745948&
0.144809&3.461721&
0.152542&3.001306
\\

\hline
NBEATS (sliding stationary) &0.220370&11.640842&0.198353&9.696634&0.181917&5.927290&0.179617&4.674210&0.170585&4.023178&0.147968&2.731574\\
NBEATS (sliding nonstationary) & 0.205605&11.160539&0.176561&7.864686&\underline{\hl{0.149617}}&5.199758&\underline{\hl{0.143902}}&4.310816&0.139380&3.995326&0.126600&2.567527\\
NBEATS (expanding stationary) & 0.267078&11.395466&0.224555&8.908530&0.204546&5.789594&0.192343&4.446875&0.187227&3.995224&0.178735&2.855251\\
NBEATS (expanding nonstationary) &\underline{\hl{0.192890}}&\underline{\hl{9.541708}}&\underline{\hl{0.171958}}&\underline{\hl{7.133679}}&0.150931&\underline{\hl{5.198079}}&0.144840&\underline{\hl{4.249010}}&\underline{\hl{0.138660}}&\underline{\hl{3.618320}}&\underline{\hl{0.121443}}&\underline{\hl{2.419592}}\\
\hline
NHITS (sliding stationary) & 0.295382&20.647861&0.264905&13.653818&0.237426&9.195409&0.219601&7.606641&0.216036&6.629158&0.192707&5.025899\\
NHITS (sliding nonstationary) &\underline{\hl{0.212958}}&\underline{\hl{10.205821}}&\underline{\hl{0.183349}}&\underline{\hl{8.264613}}&\underline{\hl{0.154622}}&\underline{\hl{5.371029}}&\underline{\hl{0.149619}}&\underline{\hl{4.187724}}&\underline{\hl{0.144124}}&\underline{\hl{3.493338}}&\underline{\hl{0.126850}}&\underline{\hl{2.366968}}\\
NHITS (expanding stationary)&0.242094&12.362057&0.230782&9.875113&0.174944&6.328451&0.170008&5.259927&0.169041&4.524931&0.152423&2.919252\\
NHITS (expanding nonstationary) & 0.223415&10.884414&0.195813&8.550162&0.169195&5.517250&0.161844&4.382561&0.157555&3.549090&0.140878&2.511572\\

\hline
DeepAR (sliding stationary) & 0.370247&16.225342&0.344032&13.691147&0.325369&\underline{\hl{7.690113}}&0.437514&6.574858&0.442863&5.773779&0.440790&5.242536\\
DeepAR (sliding nonstationary) & 0.289207&20.071034&0.258349&13.979573&0.216005&8.157852&0.201986&5.918027&0.197893&4.647609&0.173889&2.895238\\
DeepAR (expanding stationary) &\underline{\hl{0.236524}}&\underline{\hl{14.612460}}&\underline{\hl{0.204433}}&\underline{\hl{10.436315}}&\underline{\hl{0.176311}}&7.739720&\underline{\hl{0.169411}}&\underline{\hl{5.517188}}&\underline{\hl{0.161797}}&\underline{\hl{5.115320}}&\underline{\hl{0.150860}}&\underline{\hl{3.585523}}\\
DeepAR (expanding nonstationary) & 0.274395&23.638295&0.247546&16.050064&0.224100&10.881948&0.215877&8.924385&0.208023&7.082444&0.188662&4.627351\\
\hline
Vanilla Transformer (sliding stationary) &0.246883&\underline{\hl{10.738816}}&0.219728&8.794384&0.177163&5.628856&0.167761&4.457940&0.162165&4.010988&0.144691&2.681127\\
Vanilla Transformer (sliding nonstationary) &0.282478&16.684708&0.257638&10.242921&0.229150&6.796254&0.226525&5.371173&0.217671&4.708711&0.203272&3.355001\\
Vanilla Transformer (expanding stationary) &\underline{\hl{0.209274}}&15.833094&\underline{\hl{0.179388}}&10.396997&\underline{\hl{0.160312}}&6.371660&\underline{\hl{0.157623}}&5.302686&\underline{\hl{0.153099}}&5.184526&0.139460&3.222339\\
Vanilla Transformer (expanding nonstationary) &0.229673&10.819989&0.197327&\underline{\hl{7.307497}}&0.168772&\underline{\hl{5.026107}}&0.166882&\underline{\hl{3.994161}}&0.158347&\underline{\hl{3.447255}}&\underline{\hl{0.137912}}&\underline{\hl{2.335996}}\\
\hline
Informer (sliding stationary) &\underline{\hl{0.193975}}&12.408978&\underline{\hl{0.172880}}&9.540640&\underline{\hl{0.158191}}&6.047171&\underline{\hl{0.152022}}&4.740823&\underline{\hl{0.147216}}&4.043175&\underline{\hl{0.130097}}&2.732288\\

Informer (sliding nonstationary) & 0.215522&\underline{\hl{10.267083}}&0.193711&\underline{\hl{7.066434}}&0.167023&5.118081&0.161740&4.234935&0.156405&3.713271&0.140469&2.892915\\
Informer(expanding stationary)&0.214680&10.453842&0.193129&7.971823&0.168308&5.151635&0.170636&4.111550&0.165369&3.610841&0.152312&2.592824\\
Informer (expanding nonstationary) &0.230936&11.904753&0.204508&7.820154&0.179234&\underline{\hl{4.822743}}&0.162614&\underline{\hl{3.879693}}&0.159271&\underline{\hl{3.411392}}&0.145086&\underline{\hl{2.415426}}\\
\hline
PatchTST (sliding stationary) & 0.236839&17.681349&0.205936&11.336896&0.178746&6.394113&0.175206&4.806855&0.167399&3.997774&0.148863&2.740026\\
PatchTST (sliding nonstationary) & \underline{\hl{0.205970}}&10.751123&\underline{\hl{0.178671}}&7.816217&\underline{\hl{0.154022}}&5.285043&\underline{\hl{0.145980}}&4.297380&\underline{\hl{0.140401}}&3.699716&\underline{\hl{0.122231}}&2.406886\\
PatchTST (expanding stationary) & 0.238233&13.221792&0.208217&9.549650&0.188198&6.220946&0.178926&4.727653&0.171295&4.095948&0.158887&2.702615\\
PatchTST (expanding nonstationary) &0.206674&\underline{\hl{9.505235}}&0.180923&\underline{\hl{6.572556}}&0.154196&\underline{\hl{4.806834}}&0.149170&3.891382&0.143946&\underline{\hl{3.326227}}&0.127001&\underline{\hl{2.336191}}\\

\hline
TFT (sliding stationary) &
\underline{\hl{0.187052}}&\underline{\hl{9.023897}}&
\underline{\hl{0.169054}}&\underline{\hl{7.582436}}&
\underline{\hl{0.151602}}&\underline{\hl{4.861313}}&
0.14586689&\underline{\hl{3.985152}}&
\underline{\hl{0.147138}}&\underline{\hl{3.744146}}&
\underline{\hl{0.130086}}&\underline{\hl{2.539736}}\\
TFT (sliding nonstationary) &0.24939805&11.914427&
0.211641&8.224129&
0.172777&5.806524&
0.165844&5.208768&
0.156989&3.928065&
0.138504&2.675744\\

TFT (expanding stationary) &0.198127&15.236749&
0.172650&8.335560&
0.157108&6.473974&
0.154537&5.366698&
0.15127&4.30497&
0.13077044&2.83763286\\
TFT (expanding nonstationary) & 0.194463&15.672102&
0.172193&10.281637&
0.14923727&6.52532487&
\underline{\hl{0.144136}}&5.036954&
0.13860627&3.884175&
0.124012&2.687823\\
\hline
TimeGPT (sliding nonstationary) & \textbf{0.185213}&\textbf{8.673029}&
\textbf{0.160217}&\textbf{6.202305}&
\textbf{0.134315}&\textbf{4.494245}&
\textbf{0.127829}&\textbf{3.752648}&
\textbf{0.122387}&\textbf{3.236531}&
\textbf{0.109011}&\textbf{2.213337}
\\
\hline
\end{tabular}}
\end{table} 

Table \ref{tab:var2} contains the standard deviation of the forecasts from Table \ref{ml_metrics}.

\begin{table}[H]
\centering
\caption{\label{tab:var2} \small
Standard deviation of metrics for classical machine learning, traditional time series, and benchmark models for both sliding and expanding window configurations.  } 
\resizebox{\textwidth}{!}{
\begin{tabular}{l|cc|cc|cc|cc|cc|cc}
\hline & \multicolumn{2}{|c}{ DGS1 } & \multicolumn{2}{c}{ DGS2 } & \multicolumn{2}{c}{ DGS5 } & \multicolumn{2}{c}{ DGS7 } & \multicolumn{2}{c}{ DGS10 }& \multicolumn{2}{c}{ DGS30 }\\
& RMSE & MAPE & RMSE & MAPE & RMSE & MAPE & RMSE & MAPE & MAPE & RMSE & MAPE & RMSE \\
\hline  Naive & 0.176857&8.361122&
0.152006&5.974901&
0.127638&4.284059&
0.122221&3.500764&
0.117291&2.994061&
0.104616&2.0475081
\\
\hline
Naive Mean (sliding)& \underline{\hl{1.134344}}&\underline{\hl{327.476781}}&
\underline{\hl{1.053095}}&\underline{\hl{176.308774}}&
\underline{\hl{0.941467}}&\underline{\hl{71.960916}}&
\underline{\hl{0.908777}}&\underline{\hl{43.731603}}&
\underline{\hl{0.879130}}&\underline{\hl{31.322088}}&
\underline{\hl{0.815037}}&\underline{\hl{16.037434}}
 \\
Naive Mean (expanding)  &1.698148&1543.402788&
1.676391&757.577714&
1.428569&275.145696&
1.332582&169.538344&
1.234872&123.875560&
1.083507&65.867542
\\
\hline
Seasonal Naive &0.195756&9.977126&
0.166908&6.975633&
0.139491&4.818090&
0.133196&3.897796&
0.128109&3.363182&
0.113856&2.245416
 \\
\hline
Random Forests (sliding) & 0.200185 & 10.203589 &
0.174595 & 7.347237 &
0.146761 & 5.366249 &
0.140711 & 4.440794 &
0.136712& 4.060309 &
\underline{\hl{0.119932}}& 2.484318 
\\
Random Forests (expanding) &\underline{\hl{0.198143}} & \underline{\hl{8.809668}} &
\underline{\hl{0.169089}} & \underline{\hl{6.889892}} &
\underline{\hl{0.143366}} & \underline{\hl{5.151816}} &
\underline{\hl{0.137224}} & \underline{\hl{4.117087}} &
\underline{\hl{0.133940}} & \underline{\hl{3.421853}} &
0.120949& \underline{\hl{2.379350 }}
 \\
\hline
XGBoost (sliding) &\underline{\hl{0.206805}} & \underline{\hl{8.197967}} &
\underline{\hl{0.187669}} & \underline{\hl{6.358201}} &
0.169268 & \underline{\hl{5.325441 }}&
0.155966 & 4.221225 &
0.153512 &\underline{\hl{ 2.767437}} &
0.139364 &2.767437\\
XGBoost (expanding) &0.206852& 8.595066 &
0.190732 & 7.118983&
\underline{\hl{0.164261 }}& 5.430950 &
\underline{\hl{0.150865}}& \underline{\hl{4.104016 }}&
\underline{\hl{0.150212}} & 3.440415 &
0.127500 & \underline{\hl{2.439450 }}
\\
\hline
LGBM (sliding) & 0.209923 & 11.994326 &
0.180402 & 7.925223 &
0.150094 & 5.189844 &
0.144166 & 4.096100 &
0.140838 & 3.744107 &
0.127421 & 2.508858 
\\
LGBM (expanding) & \underline{\hl{0.200158 }} & \underline{\hl{8.133562}} &
\underline{\hl{0.175234}} & \underline{\hl{6.306747}} &
\underline{\hl{0.143970}} & \underline{\hl{4.633200}} &
\underline{\hl{0.138330}} & \underline{\hl{3.892541 }}&
\underline{\hl{0.134607 }}& \underline{\hl{3.224729}} &
\underline{\hl{0.119278 }}& \underline{\hl{2.257060 }}
\\ \hline

ARIMA (sliding) & \underline{\hl{0.178056}} & 8.4470 & \underline{\hl{0.151584}} & 5.9788 & \underline{\hl{0.127203}} & 4.3132 & \underline{\hl{0.122192 }}& 3.5289 & \underline{\hl{0.117364}} & 2.9945 & \underline{\hl{0.104733 }} &\underline{\hl{ 2.0471}}\\
ARIMA (expanding)&0.178253 & \underline{\hl{7.8837}} & 0.152189 & \underline{\hl{5.8641}} & 0.127280 & \underline{\hl{4.2859}} & 0.122359 &\underline{\hl{ 3.4999}} & 0.117487 & \underline{\hl{2.9849 }} & 0.104760 & 2.0495\\ \hline
DHR-ARIMA (sliding) &0.177390&    19.2421& 0.154861& \underline{\hl{6.1686}}& 0.129988&     4.4301&      0.123386&  3.6433& 0.118064&     3.1094& 0.106099&    2.1604

\\
DHR-ARIMA (expanding) & \underline{\hl{0.176259}} & \underline{\hl{9.6016}} & \underline{\hl{0.151950}} & 6.2879 & \underline{\hl{0.129006}} & \underline{\hl{4.3682}} & \underline{\hl{0.122032}} & \underline{\hl{3.5432}} & \underline{\hl{0.117069 }}&\underline{\hl{ 3.0202 }}& \underline{\hl{0.104409 }}& \underline{\hl{2.0554}}\\
\hline
diff+VAR (sliding) & \underline{\hl{0.176812}}&8.850917&
0.1524477&6.052817&
0.12985984&4.270754&
0.12449062&3.496821&
0.11968569&2.99846958&
0.10584956&2.05373023
\\
diff+VAR (expanding) &0.177444&\underline{\hl{8.243917}}&
\underline{\hl{0.152393}}&\underline{\hl{5.922836}}&
\underline{\hl{0.129550}}&\underline{\hl{4.251956}}&
\underline{\hl{0.12405484}}&\underline{\hl{3.473514}}&
\underline{\hl{0.119452}}&\underline{\hl{2.979156}}&
\underline{\hl{0.105826}}&\underline{\hl{2.045680}}
\\ \hline
VECM (sliding) & 0.208336&\underline{\hl{10.906442}}&
0.1771358&\underline{\hl{7.563252}}&
0.148699&\underline{\hl{4.943110}}&
0.14134967&\underline{\hl{3.929636}}&
0.13512554&3.414322&
0.11882189&2.33132972\\
VECM (expanding) & \underline{\hl{0.192521}}&13.046624&
\underline{\hl{0.165482}}&7.887848&
\underline{\hl{0.141066}}&5.10768405&
\underline{\hl{0.134406}}&3.98585103&
\underline{\hl{0.129452}}&\underline{\hl{3.351312}}&
\underline{\hl{0.114338}}&\underline{\hl{2.294795}}
 \\ \hline
\end{tabular}}
\end{table} 


Table \ref{tab:var3} contains the standard deviation of the forecasts from Table \ref{leaderboard}.

We can see only the traditional methods (sliding ARIMA, expanding ARIMA, expanding DHR-ARIMA,  and expanding VAR) have the lowest standard deviation. The Naive Forecast does not have the lowest standard deviation for any score, though it is effectively tied in some cases. Of the classical ML methods, expanding LGBM has the lowest standard deviation, while TimeGPT has the lowest of all ML methods.
\begin{table}[H]
\centering
\caption{\label{tab:var3} \small Standard deviation of the metrics for the final model performance.} 
\resizebox{\textwidth}{!}{
\begin{tabular}{l|cc|cc|cc|cc|cc|cc}
\hline & \multicolumn{2}{|c}{ DGS1 } & \multicolumn{2}{c}{ DGS2 } & \multicolumn{2}{c}{ DGS5 } & \multicolumn{2}{c}{ DGS7 } & \multicolumn{2}{c}{ DGS10 }& \multicolumn{2}{c}{ DGS30 }\\
& RMSE & MAPE & RMSE & MAPE & RMSE & MAPE & RMSE & MAPE& RMSE & MAPE & RMSE & MAPE\\
\hline
\hline Naive & 0.176857&8.361122&
0.152006&5.974901&
0.127638&4.284059&
0.122221&3.500764&
0.117291&2.994061&
0.104616&2.0475081
\\
\hline
Naive Mean (sliding)& 1.134344&327.476781&
1.053095&176.308774&
0.941467&71.960916&
0.908777&43.731603&
0.879130&31.322088&
0.815037&16.037434
 \\
\hline
Seasonal Naive &0.195756&9.977126&
0.166908&6.975633&
0.139491&4.818090&
0.133196&3.897796&
0.128109&3.363182&
0.113856&2.245416
 \\
\hline

ARIMA (sliding) & 0.178056 & 8.4470 & \hl{\textbf{0.151584}} & 5.9788 & \hl{\textbf{0.127203}} & 4.3132 & 0.122192 & 3.5289 & 0.117364 & 2.9945 & 0.104733 & 2.0471\\
ARIMA (expanding)&0.178253 & \hl{\textbf{7.8837}} & 0.152189 & \hl{\textbf{5.8641 }} & 0.127280 & 4.2859 & 0.122359 & 3.4999 & 0.117487 & 2.9849 & 0.104760 & 2.0495\\ \hline
DHR-ARIMA (sliding) &0.177390&    19.2421& 0.154861& 6.1686& 0.129988&     4.4301&      0.123386&  3.6433& 0.118064&     3.1094& 0.106099&    2.1604

\\

DHR-ARIMA (expanding) & \hl{\textbf{0.176259}} & 9.6016 & 0.151950 & 6.2879 & 0.129006 & 4.3682 & \hl{\textbf{0.122032}} & 3.5432 & \hl{\textbf{0.117069}} & 3.0202 & \hl{\textbf{0.104409}} & 2.0554\\
\hline
\hline
diff+VAR (expanding) &0.177444&8.243917&
0.152393&5.922836&
0.129550&\hl{\textbf{4.251956}}&
0.124054&\hl{\textbf{3.473514}}&
0.119452&\hl{\textbf{2.979156}}&
0.105826&\hl{\textbf{2.045680}}

\\ \hline

VECM (sliding) & 0.208336&10.906442&
0.177135&7.563252&
0.148699&4.943110&
0.141349&3.929636&
0.135125&3.414322&
0.118821&2.331329\\
VECM (expanding) & 0.192521&13.04662&
0.165482&7.887848&
0.141066&5.107684&
0.134406&3.985851&
0.129452&3.351312&
0.114338&2.294795\\
\hline
XGBoost (expanding) &0.206852& 8.595066 &
0.190732 & 7.118983&
0.164261 & 5.430950 &
0.150865& 4.104016 &
0.150212 & 3.440415 &
0.127500 & 2.439450 
\\
\hline
LGBM (expanding) & 0.200158 & 8.133562 &
0.175234 & 6.306747 &
0.143970 & 4.633200 &
0.138330 & 3.892541 &
0.134607 & 3.224729 &
0.119278 & 2.257060 \\ \hline

\hline\hline RNN (sliding stationary) &0.194001&9.371014&
0.162662&7.487174&
0.138587&4.900849&
0.137446&3.878842&
0.128832&3.270737&
0.110987&2.214016\\

RNN (expanding stationary) & 0.199035&8.926611&
0.165056&6.676690&
0.147118&4.587371&
0.140712&3.669374&
0.139865&3.306757&
0.126081&2.295027
\\
RNN (expanding nonstationary) &0.213878&11.242474&
0.184713&24.053278&
0.165087&10.626177&
0.156568&5.745947&
0.144808&3.461720&
0.152542&3.001306 \\
\hline

NBEATS (sliding nonstationary) & 0.205605&11.160538&0.176561&7.864685&0.149616&5.199757&0.143901&4.310816&0.139380&3.995325&0.126599&2.567526\\
NBEATS (expanding nonstationary) &0.192890&9.541707&0.171957&7.133678&0.150931&5.198078&0.144840&4.249009&0.138660&3.618319&0.121443&2.419592 \\
NHITS (sliding nonstationary)
&0.21295811&10.205821&0.1833486&8.264612&0.154622&5.371029&0.149619&4.187724&0.144124&3.493338&0.126850&2.366968\\
DeepAR (expanding stationary) & 0.236523&14.612460&0.204432&10.436314&0.176311&7.739719&0.169411&5.517187&0.161796&5.115320&0.150859&3.585523\\
\hline
Vanilla Transformer (sliding nonstationary) &0.282478&16.684707&0.257637&10.242920&0.229149&6.796253&0.226524&5.371173&0.217671&4.708710&0.203271&3.355000\\
Vanilla Transformer (expanding stationary) & 0.209273&15.833093&0.179387&10.396997&0.160312&6.371659&0.157623&5.302686&0.153099&5.184526&0.139459&3.222339\\
\hline
Informer(expanding stationary)&0.214680&10.453842&
0.193128&7.971822&
0.168308&5.151635&
0.170636&4.111549&
0.165369&3.610840&
0.152311&2.592824
\\
Informer (expanding nonstationary) &0.230935&11.904752&
0.204508&7.820154&
0.179233&4.822743&
0.162614&3.879693&
0.159271&3.411391&
0.145086&2.415425
\\
\hline
PatchTST (sliding nonstationary) & 0.205969&10.751122&
0.178671&7.816217&
0.154022&5.285042&
0.145979&4.297380&
0.140400&3.699716&
0.122231&2.406886
\\
PatchTST (expanding nonstationary) &0.206674&9.505235&
0.180923&6.572556&
0.154196&4.806834&
0.149170&3.891382&
0.143946&3.326227&
0.127001&2.336191
\\

TFT (sliding stationary) &
0.187052&9.023897&
0.169054&7.582436&
0.151602&4.861313&
0.145866&3.985152&
0.147138&3.744146&
0.130086&2.539736\\
TFT (expanding nonstationary) & 0.194463&15.672102&
0.172193&10.281637&
0.149237&6.525324&
0.144136&5.036954&
0.138606&3.884175&
0.124012&2.687823\\
\hline
TimeGPT (sliding nonstationary) & 0.185213&8.673029&
0.160217&6.202305&
0.134315&4.494245&
0.127829&3.752648&
0.122387&3.236531&
0.109011&2.213337
\\
\hline
\end{tabular}}
\end{table}

\subsection{Time Block-based Analysis}
\label{rmse_mape_plots}

In this subsection, we present some of the  results of Section \ref{results} graphically; specifically breaking down the error metrics by the hyperparameter time blocks. Figure \ref{fig:arimacomp} breaks down the sliding window ARIMA vs the expanding window ARIMA by time blocks. Figure \ref{fig:patchtstbest} compares the performance of the best configurations of PatchTST with ARIMA and the Naive Forecast, broken down by the hyperparameter time blocks.

Figures \ref{fig:arimacomp} and \ref{fig:patchtstbest} were discussed in Section \ref{results}.

In Figure \ref{fig:TFTbest}, we see that the TFT nonstationary expanding configuration also has some noticeable scores (e.g., for DGS1  in Time Block 2). It has an MAPE score of 2.452953\%, which is higher than the Naive score of 2.463761\% and ARIMA expanding score  of 2.461910\%; but these scores are effectively a tie. Here the TFT is not better than ARIMA sliding, which has a mean error of 2.418810\%. 
In Figure \ref{fig:rnn_best}, we see that the RNN also has some Time Block 3 scores that beat ARIMA but not the Naive method. However, it beats both ARIMA and Naive in Time Block 6 for most of the bonds.

\begin{figure*}
\centering
({\bf a})\\
\includegraphics[scale=0.5]{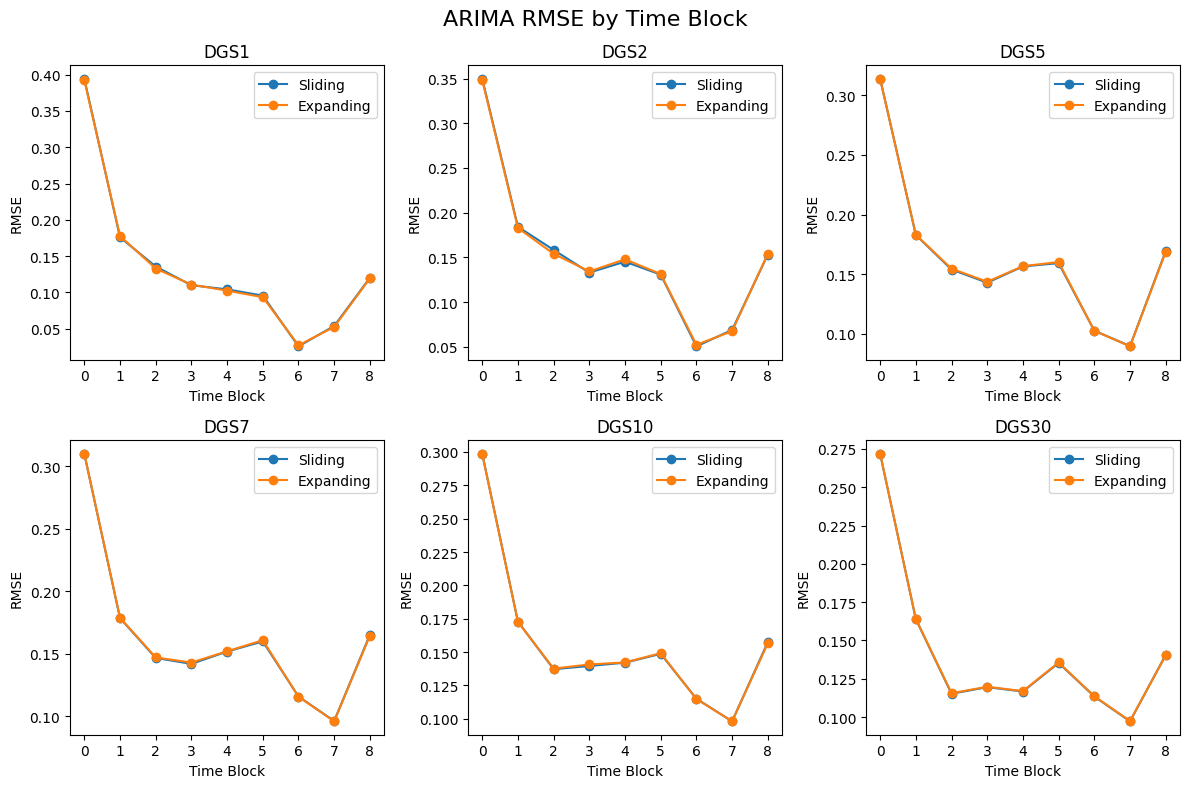}\\ 
({\bf b})\\
\includegraphics[scale=0.5]{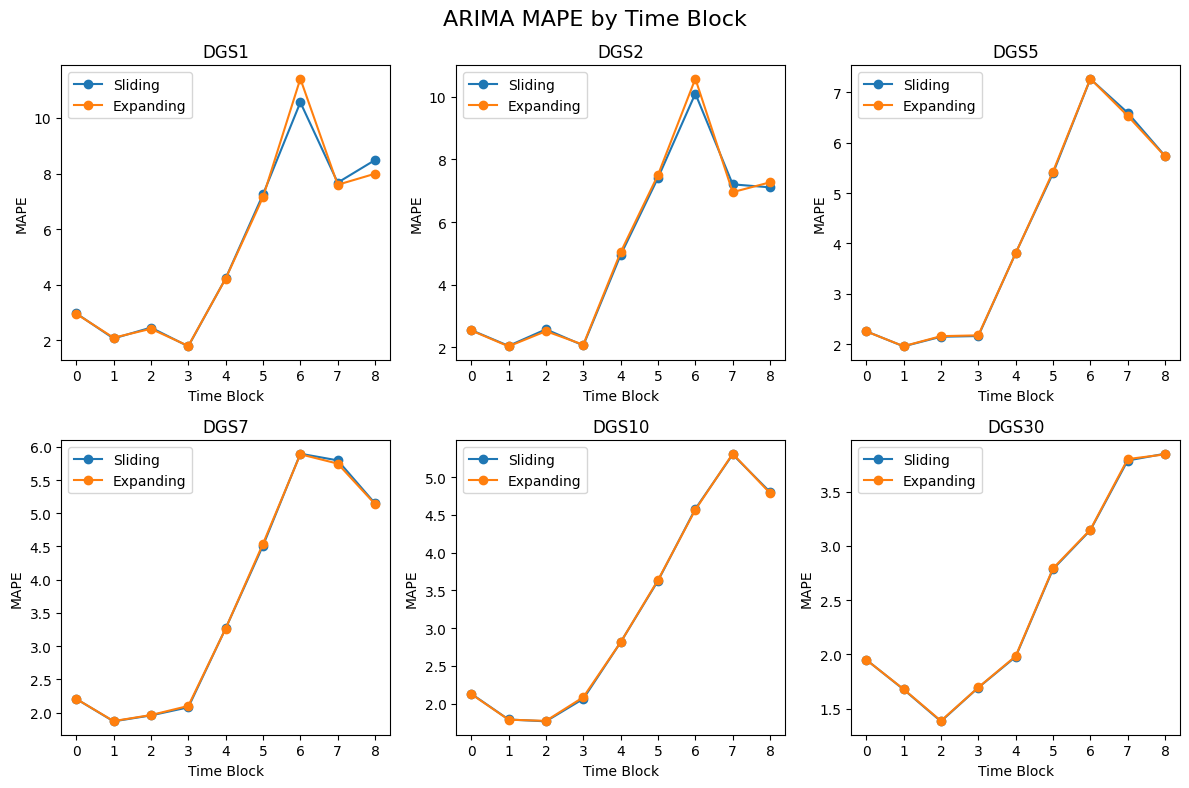}\\
\caption{RMSE metrics for ARIMA for each configuration in  all time blocks  (\textbf{b}) MAPE metrics for ARIMA for each configuration in  all time blocks.}\label{fig:arimacomp}
\end{figure*}

\begin{figure*}
\centering
({\bf a})\\
\includegraphics[scale=0.5]{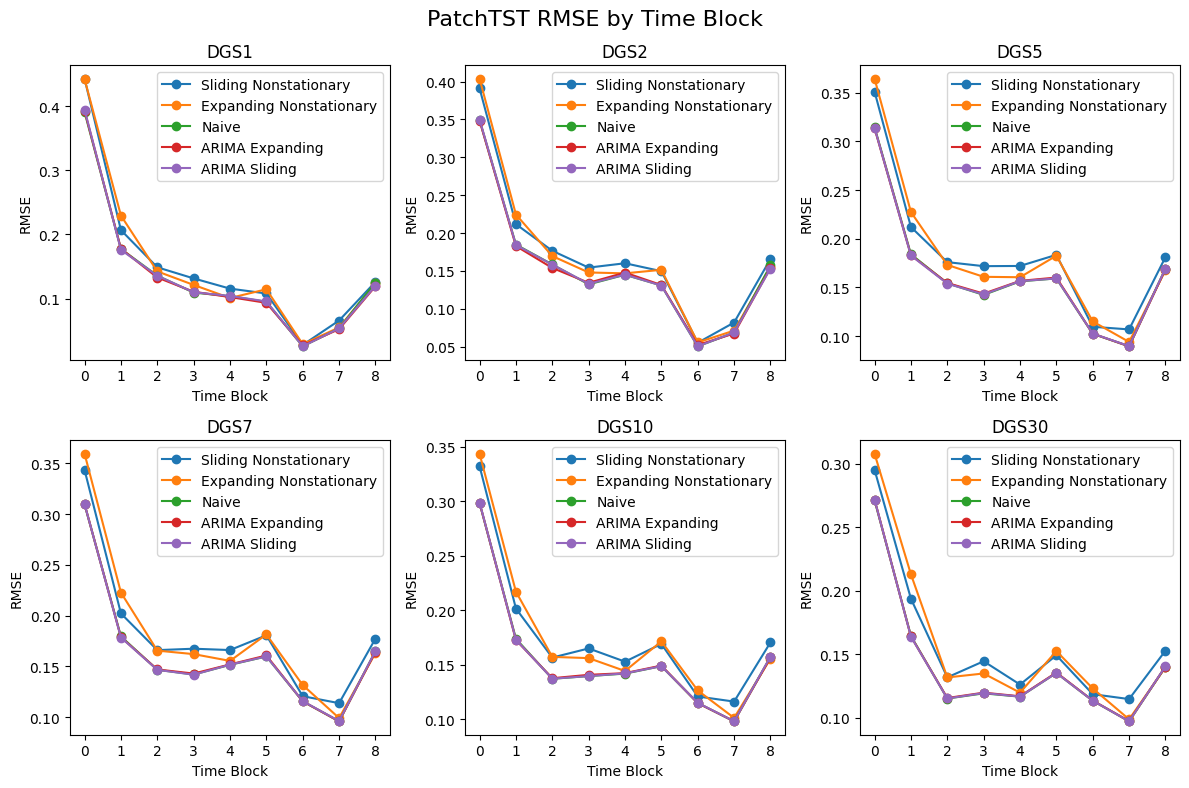}\\ 
({\bf b})\\
\includegraphics[scale=0.5]{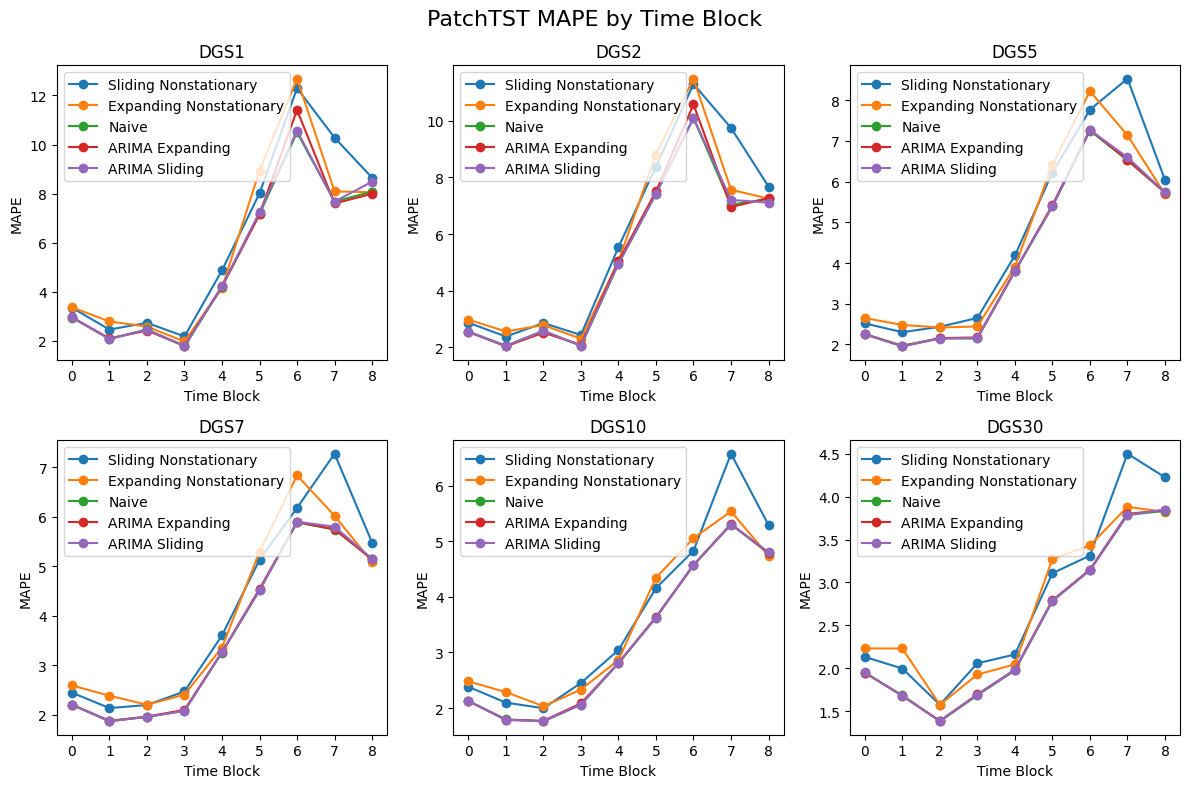}\\
\caption{(\textbf{a}) RMSE metrics for the PatchTST for the best configurations in  all time blocks, compared with the Naive Forecast. (\textbf{b}) MAPE metrics for the PatchTST  for the best configurations in  all time blocks. }\label{fig:patchtstbest}
\end{figure*}

\begin{figure*}
\centering
({\bf a})\\
\includegraphics[scale=0.5]{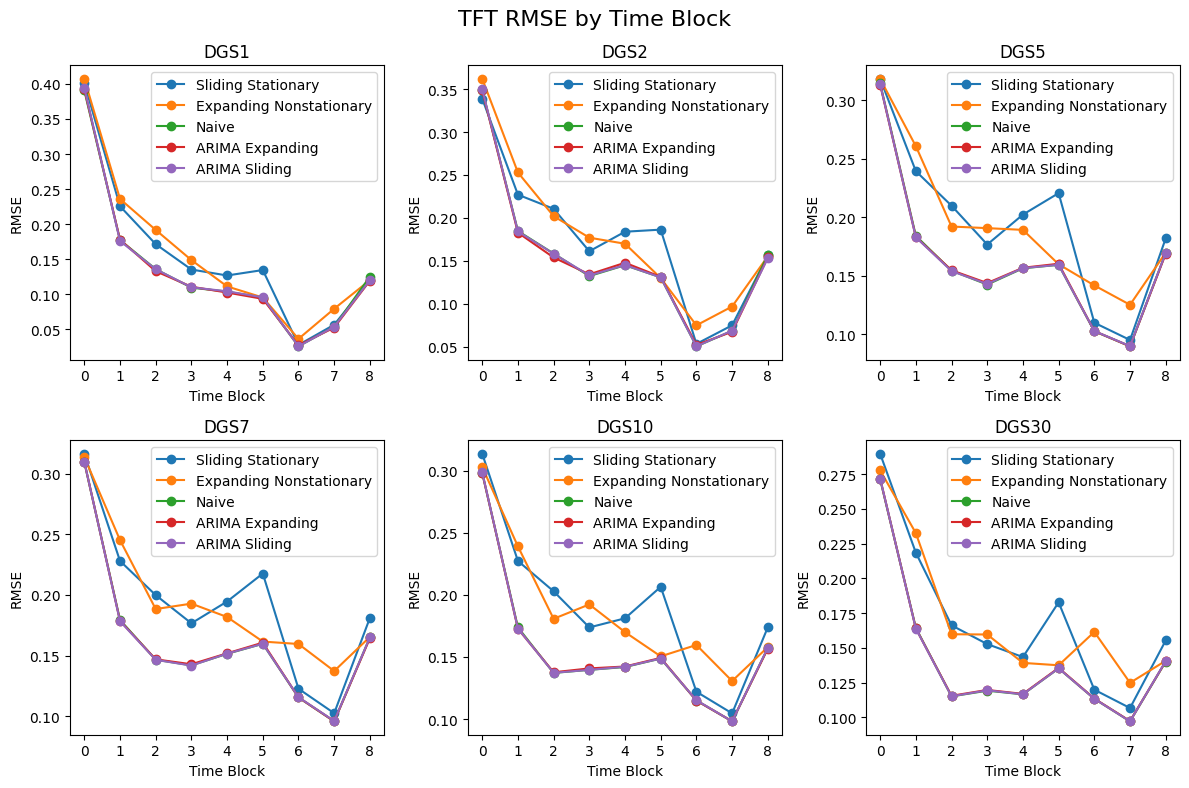}\\ 
({\bf a})\\
\includegraphics[scale=0.5]{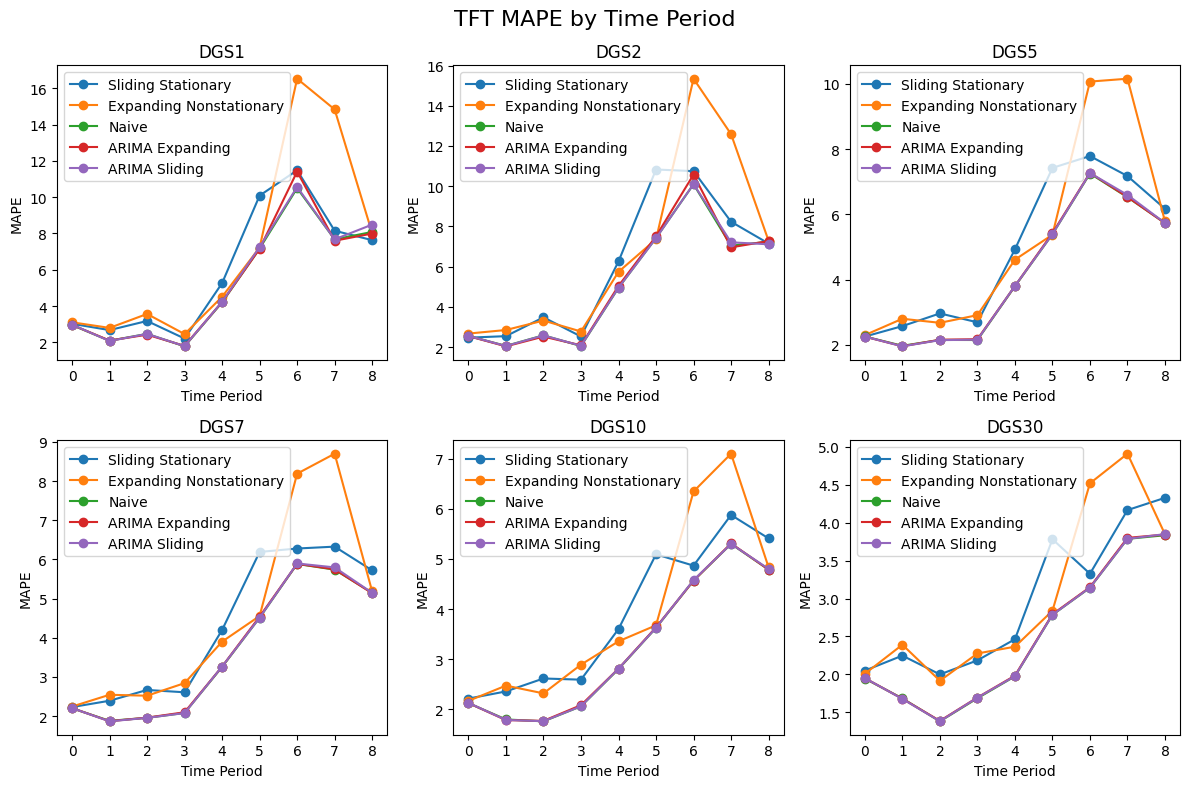}\\
\caption{(\textbf{a}) RMSE metrics for the TFT  for the best configurations in  all time blocks. (\textbf{b}) MAPE metrics for the TFT  for the best configurations in  all time blocks. }\label{fig:TFTbest}
\end{figure*}

\begin{figure}
\begin{tabular}{c}
({\bf a})\\
\includegraphics[width=6in]{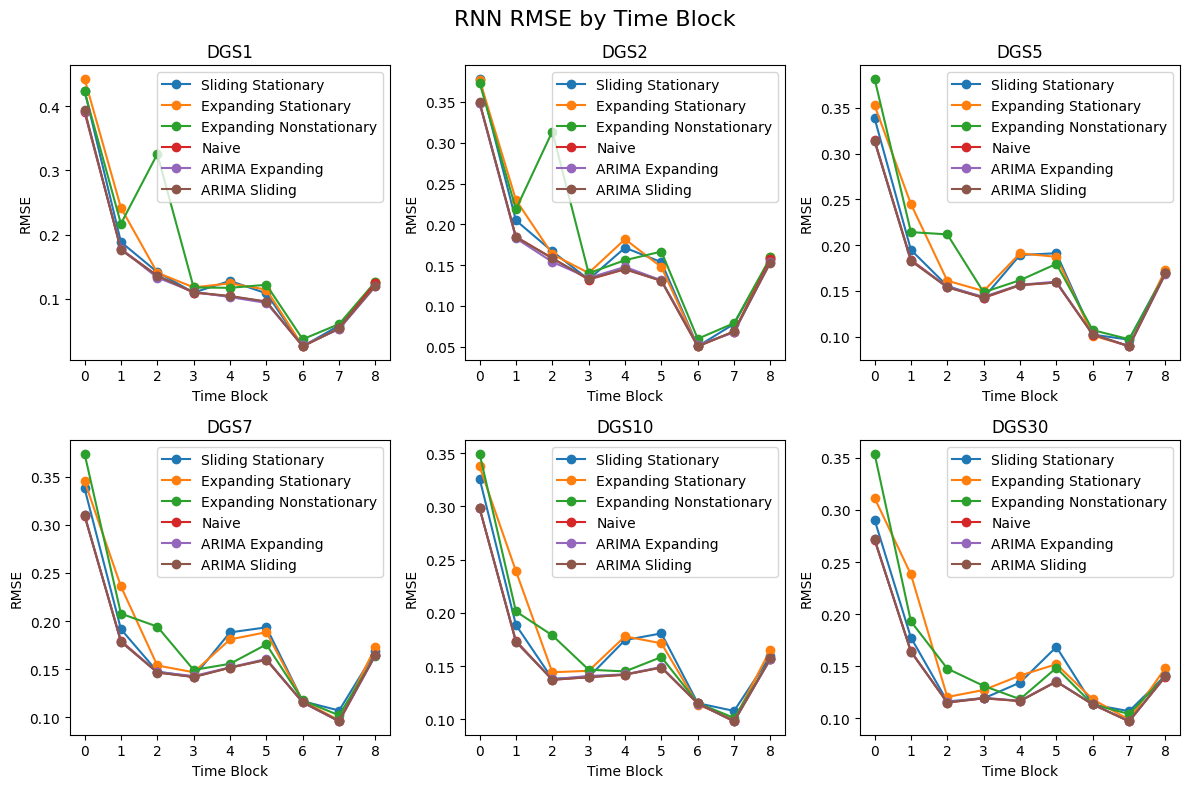}\\ 
({\bf b})\\
\includegraphics[width=6in]{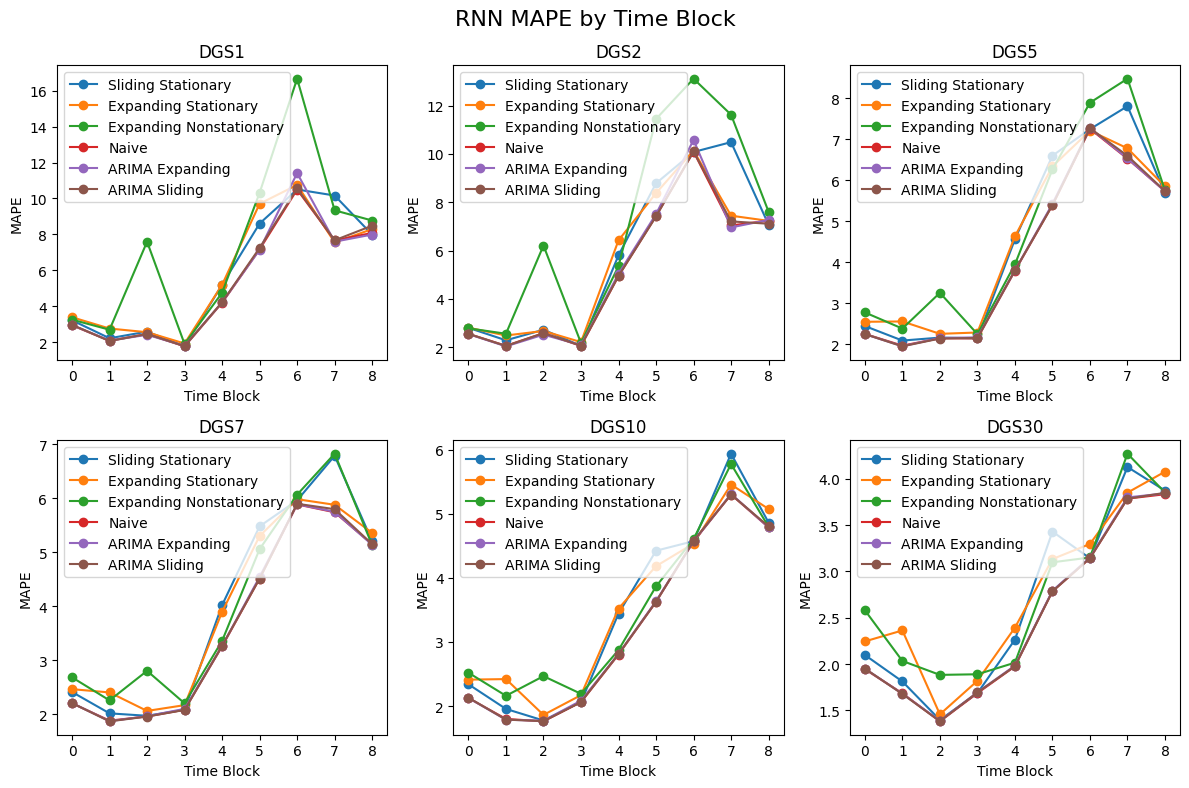}\\
\end{tabular}
\caption{ (\textbf{a}) RMSE metrics for the RNN for the best configurations in  all time blocks compared with ARIMA \& Naive.  (\textbf{b}) MAPE metrics for the RNN for the best  configurations in  all time blocks compared with ARIMA \& Naive.}\label{fig:rnn_best}
\end{figure}









\end{document}